\pdfoutput=1
\documentclass[10pt,twocolumn,letterpaper]{article}

\usepackage[pagenumbers]{cvpr}

\usepackage{xspace}
\usepackage{multirow}
\newcommand{\ours}{RealMobility}
\newcommand{\data}{RealMobility-117K}
\newcommand{\method}{RealMobility\xspace}
\newcommand{\dataset}{RealMobility-117K\xspace}

\newcommand{\numcat}{89\xspace}

\definecolor{cvprblue}{rgb}{0.21,0.49,0.74}
\usepackage[pagebackref,breaklinks,colorlinks,allcolors=cvprblue]{hyperref}

\def\confName{CVPR}
\def\confYear{2027}

\title{RealMobility-117K: A Realistic Articulated Object Dataset via Aligning Video Motion with 3D Generation}

\author{
Puxin Yan$^{1,2,3,4}$ \quad Hui-Ze Tan$^{2,5}$ \quad Yuanzheng Zhao$^{7}$ \quad Liu Liu$^{4}$ \quad Hongyi Chen$^{1,6}$ \\
Badi Li$^{2}$ \quad Ruicheng Zhang$^{8}$ \quad Sicheng Yu$^{9,4}$ \quad Zhizhong Su$^{4}$ \quad Wei Sui$^{3}$ \\
Yi Ma$^{1,2,5}$ \quad Yuan Liu$^{6}$ \quad Yizhou Yu$^{1,2}$ \quad Shenghua Gao$^{1,2,5,\dagger}$ \\[1ex]
{\small $^{1}$Shenzhen Loop Area Institute \quad $^{2}$The University of Hong Kong \quad $^{3}$D-Robotics \quad $^{4}$Horizon Robotics} \\
{\small $^{5}$TranscEngram \quad $^{6}$The Hong Kong University of Science and Technology \quad $^{7}$Sun Yat-sen University} \\
{\small $^{8}$Tsinghua University \quad $^{9}$The Hong Kong University of Science and Technology (Guangzhou)} \\[0.5ex]
{\small $^{\dagger}$Corresponding author}
}

\begin{document}
\twocolumn[{%
\renewcommand\twocolumn[1][]{#1}%
\maketitle
\begin{center}
  \includegraphics[width=0.95\textwidth]{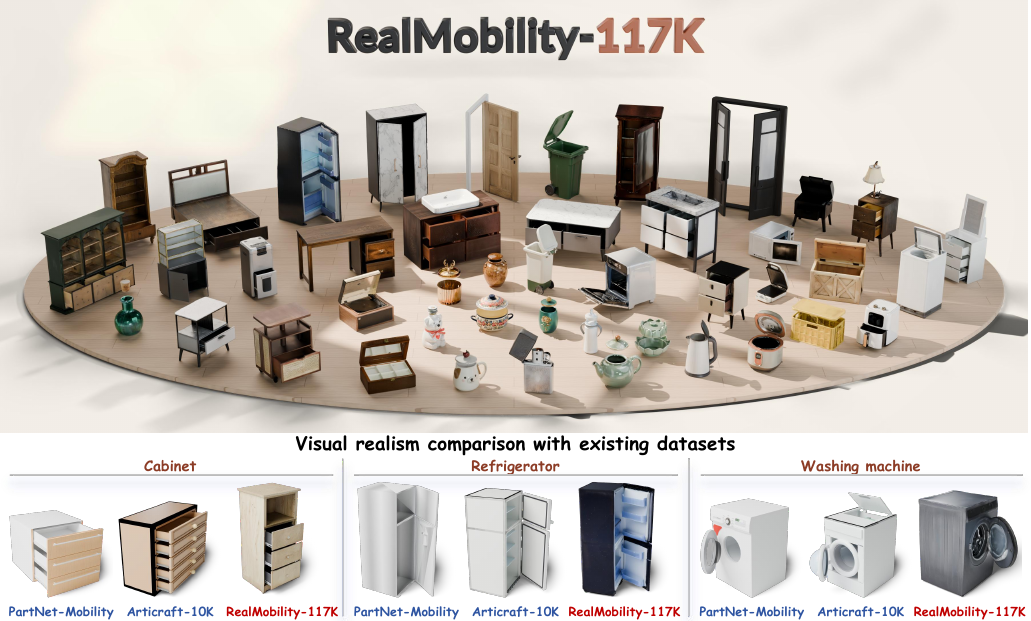}
  \captionof{figure}{\textbf{\dataset.} Top: articulated assets from \dataset. Bottom: visual comparison with PartNet-Mobility and Articraft-10K on the same categories.}
  \label{fig:teaser}
\end{center}%
}]

\begin{abstract}
Articulated objects are important for games, virtual reality, and robotic interaction, which require assets with both correct motion structures and realistic appearance. Existing methods struggle to achieve both at scale: manual modeling is costly, learning-based generation is limited by scarce training data, and agent-based procedural modeling lacks realistic textures and details. Our key insight is that video generators can show how parts move, while 3D generators can produce detailed geometry and textures. We thus propose RealMobility, a training-free framework that aligns video motion with 3D generation: a pixel-aligned 3D generator reconstructs keyframes of a generated motion video as sparse voxels, where changes across states reveal movable parts and joints, and then decodes detailed textured meshes. With RealMobility, we build RealMobility-117K, 117,130 articulated assets from 89 everyday categories, the largest human-verified articulated object dataset to date, with structural, semantic, and physical property annotations. Experiments show that RealMobility outperforms single-image articulated generation methods on ACD-135 across all metrics and receives 76.9\% and 79.7\% of the votes for input fidelity and texture quality in a user study. Particulate trained on about 10K of our assets reaches a rest-state gIoU of 0.452 on ACD-135, higher than 0.295 and 0.171 when trained on PartNet-Mobility and Articraft-10K.

\end{abstract}

\section{Introduction}
\label{sec:intro}

Articulated objects are common in everyday environments and are important for games, virtual reality, and robotic interaction.
These applications require assets not only to have correct motion structures, but also to look like real objects: immersive virtual environments rely on realistic geometry and textures, and robot models trained in simulation also need realistic object appearance to reduce the visual gap between simulation and the real world~\cite{artvip,tobin2017domain}.

However, existing construction approaches struggle to achieve both realism and scale.
Manually modeling articulated assets~\cite{sapien,artvip,artiverse} is costly and hard to scale.
Learning-based 3D generation methods~\cite{nap,cage,singapo,pact} automatically generate articulated assets by learning object shapes, appearances, and part relationships, but they are limited by scarce articulated training data and generalize poorly to unseen objects.
Recently, Articraft~\cite{articraft} and USDCraft~\cite{usdcraft} use agents to automatically write modeling programs, greatly improving scalability.
However, procedural modeling struggles to reproduce the rich textures and fine local structures of real objects, so the resulting assets still have limited visual realism.
This leads to our central question: how can we build visually realistic articulated assets at scale while keeping their motion structures reliable?

Our key insight is that pretrained 3D generators can already produce detailed geometry and rich textures~\cite{trellis,cupid,hunyuan3d2}, while video generators contain rich world knowledge and can show how articulated parts move~\cite{wan,cogvideox}.
A natural idea is to combine the two: by connecting motion with geometry, we can turn motion cues in videos into 3D parts and joint structures, making assets movable while preserving their geometric and texture details.

To this end, we propose \method, a training-free framework for articulated asset generation.
Given an image, \method first generates a video of part motion and uses a pixel-aligned 3D generator~\cite{cupid} to establish correspondences between keyframes and sparse voxels.
We then identify movable parts from geometric changes across states and map tracking correspondences from the video~\cite{cowtracker,cotracker} to the 3D positions of these parts to estimate joint parameters.
Based on this structure, we design a rule-based algorithm to complete missing interior part structures.
Finally, the sparse voxels are decoded into detailed textured meshes, producing articulated assets with both visual realism and movability.

\begin{table*}[t]
  \definecolor{tabgreen}{RGB}{34,139,69}%
  \definecolor{tabred}{RGB}{200,50,50}%
  \definecolor{tabamber}{RGB}{225,150,20}%
  \def\cmark{\textcolor{tabgreen}{$\checkmark$}}%
  \def\xmark{\textcolor{tabred}{$\boldsymbol{\times}$}}%
  \def\pmark{\textcolor{tabamber}{$\triangle$}}%
  \def\nr{\textcolor{black!45}{--}}%
  \def\hd#1{\begin{tabular}[b]{@{}c@{}}#1\end{tabular}}%
  \centering
  \footnotesize
  \setlength{\tabcolsep}{3.5pt}
  \renewcommand{\arraystretch}{1.15}
  \resizebox{\linewidth}{!}{%
  \begin{tabular}{@{}llccccccccccrr@{}}
    \toprule
    & & &
    \multicolumn{3}{c}{Structure \& semantics} &
    \multicolumn{6}{c}{Physical properties} & & \\
    \cmidrule(lr){4-6} \cmidrule(lr){7-12}
    Dataset & \hd{Geometry\\source} & \hd{Human\\verified} &
    Joints & \hd{Part\\names} & \hd{Function\\text} &
    \hd{Metric\\scale} & Material & Mass & Friction & \hd{Joint\\damping} & \hd{Collision\\geometry} &
    \#Objects & \#Cat. \\
    \midrule
    PartNet-Mobility~\cite{sapien}             & Manual modeling & \cmark & \cmark & \cmark & \xmark & \xmark & \xmark & \xmark & \xmark & \xmark & \xmark & 2,346 & 46 \\
    ArtVIP~\cite{artvip}                       & Manual modeling & \cmark & \cmark & \cmark & \xmark & \cmark & \xmark & \pmark & \pmark & \cmark & \pmark & 992     & 37 \\
    PhysX-Mobility~\cite{physxanything}        & Manual modeling & \cmark & \cmark & \cmark & \cmark & \cmark & \cmark & \cmark & \xmark & \xmark & \xmark & 2,024   & 47 \\
    Artiverse~\cite{artiverse}                 & Manual modeling & \cmark & \cmark & \cmark & \xmark & \cmark & \cmark & \cmark & \xmark & \xmark & \pmark & 5,402   & 88 \\
    \midrule
    Infinigen-Articulated~\cite{infinigensim}  & Procedural      & \xmark & \cmark & \cmark & \xmark & \cmark & \xmark & \cmark & \cmark & \cmark & \cmark & 8,226   & 17 \\
    Articraft-10K~\cite{articraft}             & Procedural      & \cmark & \cmark & \cmark & \xmark & \cmark & \xmark & \pmark & \xmark & \pmark & \xmark & 10,018  & 245 \\
    USDCraft-10K~\cite{usdcraft}               & Procedural      & \xmark & \cmark & \cmark & \xmark & \cmark & \cmark & \cmark & \cmark & \cmark & \cmark & 10,000  & 500+ \\
    RFW-A-400K~\cite{rfwa}                     & Procedural      & \xmark & \cmark & \cmark & \cmark & \cmark & \cmark & \cmark & \cmark & \pmark & \xmark & 402,410 & 531 \\
    \midrule
    \textbf{\dataset (ours)}                   & \textbf{3D generation} & \cmark & \cmark & \cmark & \cmark & \cmark & \cmark & \cmark & \cmark & \cmark & \cmark & \textbf{117,130} & \textbf{89} \\
    \bottomrule
  \end{tabular}%
  }
  \caption{\textbf{Comparison with existing articulated object datasets.}
  \emph{Procedural}: rule-based generators or LLM-written modeling programs.
  \emph{Human verified}: every asset is modeled, annotated, or inspected by humans.
  \emph{Function text}: per-part function description.
  \emph{Material}: per-part material category.
  \emph{Mass}: per-part mass or density.
  \emph{Friction}: surface friction.
  \emph{Collision geometry}: simplified colliders (convex pieces or primitives), not the visual mesh.
  \#Objects follow each paper, or the public release when no exact count is given.
  \pmark: only a subset of assets.
  \dataset is the largest human-verified dataset and the only one with all annotations.}
  \label{tab:dataset_comparison}
\end{table*}

Using \method, we build \dataset (Fig.~\ref{fig:teaser}), a more visually realistic articulated object dataset with 117,130 objects from \numcat everyday categories.
We manually inspect every object in \dataset to ensure that it has a correct motion structure as well as detailed geometry and textures.

To support downstream learning and simulation, we further provide two types of annotations for each object: structural and semantic annotations, and physical property annotations.
Structural and semantic annotations include the motion structure, part names, and part function descriptions.
Physical property annotations include the real-world scale of the object, the material, mass, and friction and restitution coefficients of each part, and the damping and friction of each joint.
Based on these annotations, we build collision geometry for each part, compute centers of mass and inertia, and provide conversion tools for one-click import into mainstream simulators such as MuJoCo~\cite{mujoco}, Genesis~\cite{genesis}, SAPIEN~\cite{sapien}, and Isaac Sim~\cite{isaacsim}.
As shown in Table~\ref{tab:dataset_comparison}, \dataset is the largest articulated object dataset in which every object is verified by humans, and the only one that provides all of these annotations.

We conduct extensive experiments on both the method and the dataset.
For the method, \method outperforms all eight single-image articulated generation methods on ACD-135~\cite{s2o}.
In a user study covering 25 categories, assets generated by \method are preferred over those of two agent-based modeling methods in 76.9\% of votes for input fidelity and 79.7\% for texture realism.
For the dataset, we use Particulate~\cite{particulate} as the model and compare different sources of training data.
At the same training scale as Articraft-10K, the model trained on \dataset achieves a rest-state gIoU of 0.452 on ACD-135, substantially higher than 0.171 with Articraft-10K and 0.295 with PartNet-Mobility.
This shows that \dataset provides more effective supervision for learning articulated structures.
We also demonstrate the use of \dataset in robot simulation.

Our main contributions are as follows:
\begin{itemize}
  \item We propose \method, which turns motion cues in videos into 3D parts and joint structures and generates visually realistic articulated assets without training.
  \item We build \dataset, a visually realistic articulated asset dataset with 117,130 objects, providing rich textures, detailed geometry, human-verified motion structures, and rich semantic and physical property annotations.
  \item We conduct experiments showing that \method outperforms single-image articulated generation methods in motion structure recovery and agent-based methods in visual quality, and that \dataset provides more effective supervision for articulation learning than existing datasets and can be used in robot simulation.
\end{itemize}
Our code and data are available at \url{https://github.com/yypxyypx/RealMobility-117K}.

\section{Related Work}
\label{sec:related}

\begin{figure*}[t]
  \centering
  \includegraphics[width=\textwidth]{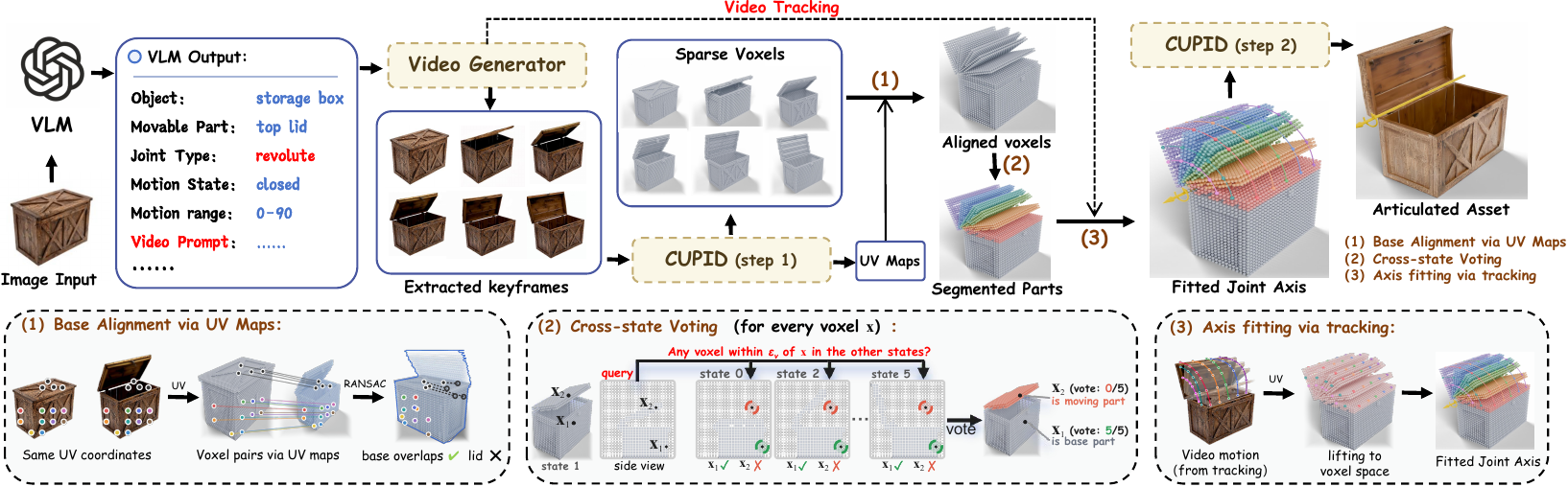}
  \caption{\textbf{Overview of \method.} A VLM recognizes the joints and prompts a video generator. CUPID reconstructs the keyframes as sparse voxels with UV maps, which are (1) aligned via UV maps, (2) segmented by cross-state voting, and (3) used to fit joint axes from video tracking. The second step of CUPID then decodes the articulated asset.}
  \label{fig:pipeline}
\end{figure*}

\paragraph{Articulated asset construction.}
Constructing articulated assets requires obtaining both part geometry and motion relationships.
Learning-based methods learn this connection from existing articulated data and generate parts and joints from inputs such as images~\cite{nap,cage,singapo,pact,urdfanythingplus,monoart,pwmartgen,physxanything,physxomni,urdformer,dipo}.
For example, SINGAPO~\cite{singapo} first predicts a part connectivity graph and then retrieves part geometry, and PAct~\cite{pact} generates part geometry with a 3D generator.
However, limited by scarce articulated training data, these methods generalize poorly to unseen objects.
Articraft~\cite{articraft} and related methods~\cite{articulateanything,real2code,usdcraft,rfwa} instead use agents to write modeling programs that explicitly define geometry, parts, and joints, which makes the motion structure easy to control and check.
However, programmatic representations struggle to reproduce rich textures and fine local structures.
\method does not rely on articulated training data: it obtains geometry and appearance from pretrained 3D generators and recovers parts and joints from generated videos, combining visual realism with movability.

\paragraph{Articulated object datasets.}
Different construction approaches have produced articulated datasets that differ in scale, visual quality, and annotations.
PartNet-Mobility~\cite{sapien}, Shape2Motion~\cite{shape2motion}, and GAPartNet~\cite{gapartnet} provide widely used part and motion annotations, while ArtVIP~\cite{artvip} and Artiverse~\cite{artiverse} further focus on visual quality, interior structures, or physical properties, extending the use of these assets in fine-grained understanding and interaction tasks.
With the development of automatic construction methods~\cite{infinitemobility}, datasets such as Articraft-10K~\cite{articraft}, USDCraft-10K~\cite{usdcraft}, and RFW-A~\cite{rfwa} have further expanded the scale of available articulated assets.
However, manually curated and annotated datasets are hard to keep scaling, while large procedurally constructed collections still lack rich textures and local geometric details, which limits their visual realism.
\dataset uses generative priors to provide richer textures and finer local geometry, improving the visual realism of articulated assets.
It is the largest articulated object dataset in which every object is verified by humans, and it provides structural, semantic, and physical property annotations to support articulation learning and robot simulation.

\paragraph{Articulation recovery from generative priors.}
Pretrained video and 3D generative models provide motion and appearance priors for constructing articulated assets~\cite{dragapart,puppetmaster}.
DreamArt~\cite{dreamart} combines generated videos with differentiable rendering optimization to recover motion structure, Rest2Art~\cite{rest2art} associates 3D anchors on a rest-state mesh with video tracks to estimate joints, and FreeArt3D~\cite{freeart3d} jointly optimizes geometry and motion parameters through score distillation.
In contrast, \method uses the pixel--geometry correspondences provided by CUPID~\cite{cupid} to lift part motion from generated videos directly into its sparse voxels, recovers parts and joints there, and then decodes the completed structure into a detailed asset, so that structure recovery and detail generation are performed in a single pipeline.

\section{RealMobility: Articulated Asset Generation}
\label{sec:method}

\method takes a single object image as input and generates an articulated asset in three steps (Fig.~\ref{fig:pipeline}): joint recognition and motion video generation (\cref{sec:motion_synthesis}), part segmentation and axis fitting (\cref{sec:alignment_segmentation}), and interior completion and detailed asset generation (\cref{sec:completion_generation}).

\subsection{Joint Recognition and Motion Generation}
\label{sec:motion_synthesis}

\paragraph{Recognizing joints and generating motion.}
We consider three joint types: revolute, prismatic, and free.
Given an input image $I$, a vision-language model (VLM) identifies all joints of the object with a recognition prompt $\mathcal P_{\mathrm{rec}}$:
\begin{equation}
\mathcal A=\operatorname{VLM}(I,\mathcal P_{\mathrm{rec}})
=\{(\ell_k,j_k,s_k,r_k)\}_{k=1}^{M},
\label{eq:vlm_recognition}
\end{equation}
where, for the $k$-th of $M$ joints, $\ell_k$ describes the name and location of its part, $j_k$ is the joint type, $s_k\in\{\text{closed},\text{open}\}$ is the initial state, and $r_k$ is the motion range (e.g., 0--90$^\circ$).
The motion range is directly used as the open limit of the joint, and the other cues decide which parts move and how; 3D joint parameters such as the axis are estimated by the later geometric steps.

We group the joints by type and generate one motion video per group with a pretrained video model, MiniMax H3~\cite{minimaxh3}.
The groups are processed as chained stages: the last frame of one stage is the input image of the next.
If a joint is initially closed, the video shows it opening; if it is initially open, the video shows it closing.
The prompt keeps the camera, the base (the static part of the object), and non-target parts still, so that changes in each video come only from the target joints of the stage.

\paragraph{Selecting keyframes.}
Generated motion often pauses or changes speed, so we sample frames by motion progress instead of time.
CoWTracker~\cite{cowtracker} estimates the overall pixel displacement between adjacent frames, whose normalized cumulative sum gives the motion progress of each frame; we take six evenly spaced progress values in $[0,1]$ and use the first frames that reach them as keyframes.

\subsection{Part Segmentation and Axis Fitting}
\label{sec:alignment_segmentation}

\paragraph{Per-state voxel reconstruction.}
We order the six keyframes of a stage by openness and denote them as states 0 to 5, where state 0 is closed and state 5 is the most open; for videos with closing motion, the order is reversed.
Each state is reconstructed by the first step of the pixel-aligned 3D generator CUPID~\cite{cupid} into a sparse voxel set $V_t$, in which each voxel $\mathbf x$ carries its UV coordinate $\pi_t(\mathbf x)$ on the image of state $t$.
The following steps operate in voxel space within each stage, and their results are finally merged onto a reference state.

\paragraph{Base alignment via UV maps.}
Independently reconstructed voxel sets differ in scale, position, and orientation.
We align all states of a stage to state 0, so that the base coincides across states while movable parts lie at different positions.
Since the camera and base stay still in the video, the same UV coordinate on the base corresponds to the same surface point in every state.
We therefore sample a set of UV coordinates $\mathcal U$, take the voxels $\mathbf x_t(u)\in V_t$ and $\mathbf x_0(u)\in V_0$ with UV coordinate $u$ as correspondences, and use RANSAC~\cite{ransac} to find the similarity transform $T(\mathbf x)=sR\mathbf x+\mathbf b$ that aligns the most of them:
\begin{equation}
T_t^\star=\arg\max_{T}\sum_{u\in\mathcal U}
\mathbb{I}\!\left[\|T(\mathbf x_t(u))-\mathbf x_0(u)\|_2<\varepsilon_a\right],
\label{eq:state_alignment}
\end{equation}
where $\mathbb{I}[\cdot]$ is the indicator function.
Under the strict threshold $\varepsilon_a$, correspondences on movable parts or occluded regions are not counted, so $T_t^\star$ is determined by the base.
We denote the aligned voxels as $\widetilde V_t=T_t^\star(V_t)$ and discard states with too few aligned correspondences.

\paragraph{Cross-state voting and part segmentation.}
After alignment, base voxels appear at the same position in most states, whereas voxels of movable parts do not.
For each valid state $t$ and each voxel $\mathbf x\in\widetilde V_t$, we count the other states that have a voxel nearby:
\begin{equation}
c_t(\mathbf x)=\sum_{\substack{t'\in\mathcal S\\t'\ne t}}
\mathbb{I}\!\left[d(\mathbf x,\widetilde V_{t'})<\varepsilon_v\right],
\label{eq:state_support}
\end{equation}
where $\mathcal S$ is the set of valid states and $d$ is the distance to the nearest voxel.
Voxels with fewer than $\eta$ votes become movable-part candidates.
We cluster the candidates and keep the $n$ largest clusters as movable parts, where $n$ is the number of joints in the stage given by the VLM; the remaining voxels belong to the base.
When $n>1$, clusters are matched one-to-one across states by their contact locations with the base, which stay nearly fixed during motion.

\paragraph{Axis fitting.}
\label{sec:joint_estimation}
For revolute joints, CoWTracker provides the UV coordinates of tracked surface points $p$ in each keyframe, and looking them up within the segmented part gives their aligned 3D positions $\mathbf x_t^p$.
Since the rotation axis is shared by all states while the angle varies, we solve
\begin{equation}
\min_{\mathbf a,\mathbf c,\{\theta_{tt'}\}}\ \sum_{t,t'}\sum_{p}
\left\|R(\mathbf a,\theta_{tt'})(\mathbf x_t^p-\mathbf c)+\mathbf c-\mathbf x_{t'}^p\right\|_2^2,
\label{eq:shared_axis}
\end{equation}
where $\mathbf a$ is the axis direction, $\mathbf c$ is a point on the axis, $\theta_{tt'}$ is the rotation angle between states $t$ and $t'$, and unreliable tracks are removed beforehand.
For flat parts such as door panels, the axis can also be obtained more cheaply as the common intersection of the part planes across states; the system chooses between the two ways automatically to support large-scale construction (details in Appendix~\ref{sec:joint_fitting}).
After obtaining the axis, we use the fitted angle between states 5 and 0 to rotate the part back to state 0, which defines the zero (closed) position of the joint, and take the motion range $r_k$ from the VLM as the open limit.
For prismatic joints, the sliding direction follows the part displacement between states 0 and 5, the zero position is the closed position, and the travel is set by a fixed pull-out ratio of the drawer length after interior completion (\cref{sec:completion_generation}).
For free parts, we only record the parent part and put the part back at its closed position; free parts have no motion axis or motion range.

\paragraph{Merging onto the reference state.}
The final asset is generated in a reference state, namely state 5 of the last stage, where inner surfaces and cavities are most exposed and free parts are lifted off.
For mixed-joint objects with multi-stage motion videos, we align all stages to the reference state and label its voxels by voting stage by stage in reverse order, starting from the last stage; voxels that are already labeled do not take part in later votes.
Since the stages are chained, parts moved in earlier stages remain at their moved positions in the reference state and are still identified.
The joint parameters of all stages are transformed into the reference state accordingly.

\subsection{Interior Completion and Asset Generation}
\label{sec:completion_generation}

\paragraph{Completing interior structures.}
Although drawers in the reference state are pulled out, the generated result often contains only the front panel and incomplete side, bottom, and back panels; the inside of the base is also often filled with solid voxels, leaving no space for the drawer to slide in, so decoding it directly would give drawers that cannot open or close properly.
We therefore complete the drawers with rules in voxel space: taking the front panel as the cross-section, we complete the drawer box inward along the sliding direction up to 0.95 of the base depth and carve the corresponding cavity in the base.

\paragraph{Generating detailed geometry and textures.}
Conditioned on the keyframe image of the reference state, the second step of CUPID generates sparse latent features for the whole object at the completed voxels, which are split by part labels and decoded into textured meshes.
Sharing one set of latent features keeps the details consistent and the seams aligned, while each part remains a separate mesh driven by its joint parameters.

\section{RealMobility-117K: Dataset Building}
\label{sec:dataset}

\begin{figure*}[t]
  \centering
  \includegraphics[width=\textwidth]{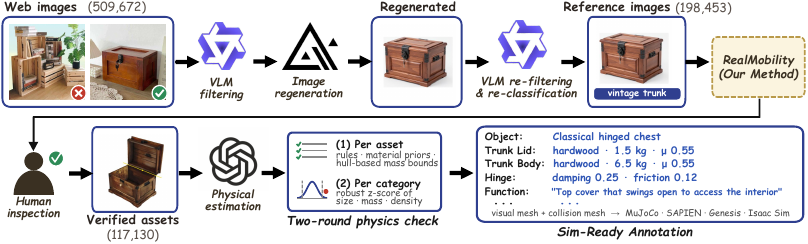}
  \caption{\textbf{Construction of \dataset.} Web images are filtered and regenerated into clean reference images, from which \method generates assets that are manually inspected. A VLM then estimates physical properties, which pass two rounds of checks to form sim-ready annotations. Numbers are the counts kept at each step.}
  \label{fig:construction}
\end{figure*}

We build \dataset in two stages (Fig.~\ref{fig:construction}): turning 509,672 web images into 117,130 human-verified articulated assets, and adding sim-ready annotations to each asset.

\paragraph{Reference image preparation.}
We collect 509,672 object images from the public internet.
These images often contain cluttered backgrounds, multiple objects, or incomplete views, so we first filter them with a vision-language model (VLM), Qwen~\cite{qwen3vl}, keeping images with a clear category, an identifiable target object, and movable parts.
For each kept image, the VLM extracts a description of the object's shape and appearance and rewrites it into a generation prompt, and FLUX~\cite{flux} then regenerates an image with a clean background and a complete object.
The VLM then filters and re-classifies the regenerated images, checking the category, structural completeness, and visual quality, which yields 198,453 reference images.

\paragraph{Asset generation and human inspection.}
We feed the reference images into \method to generate articulated assets and manually inspect every asset.
The inspection covers three aspects: the motion structure, including part segmentation, the position and direction of joint axes, and whether the opening and closing are plausible; the overall geometry, including overall proportions, the quality of interior structures, floating fragments, and residual geometry on the back of the object; and the textures, including their completeness and visual quality.
Assets with defects in geometry, interior structures, or textures are discarded rather than repaired, leaving 117,130 verified assets.
Manual edits are limited to refining joint parameters, namely the axes and part segmentation of revolute joints and the closed positions of movable parts (Appendix~\ref{sec:inspection}).

\paragraph{Structural and semantic annotations.}
The motion structure is generated by \method and checked during human inspection, where the joint parameters are refined as described above when needed.
It records the connections between parts, the type of each joint, and the axis and motion range of each revolute or prismatic joint.
On top of this, GPT-6 Astra~\cite{gpt6astra} generates part names and function descriptions from rendered images of the asset; for example, the lid in Fig.~\ref{fig:construction} is described as ``top cover that swings open to access the interior.''

\paragraph{Physical property annotations.}
The physical properties include the real-world scale of the object, the material, mass, friction coefficient, and restitution coefficient of each part, and the damping and friction of each joint.
GPT-6 Astra estimates them from multi-view renderings of the whole object and renderings that highlight each part.
The estimates must pass two rounds of checks: per-asset checks based on rules, material priors, and convex-hull-based mass bounds, and per-category checks that compute robust z-scores of size, mass, and density within each category to find outliers.
Assets that fail the checks are sent back to GPT-6 Astra with the specific reasons for re-estimation.
Based on the verified properties, we build collision geometry for each part with CoACD~\cite{coacd} convex decomposition and compute centers of mass and inertia from the masses.
We also provide conversion tools that import the visual and collision meshes into mainstream simulators such as MuJoCo, SAPIEN, Genesis, and Isaac Sim with one click.

\paragraph{Dataset statistics.}
\dataset contains 117,130 articulated assets from \numcat common everyday categories, covering furniture, household appliances, containers, and more.
The dataset contains 216,508 joints, including 72,290 revolute joints (33.4\%), 125,007 prismatic joints (57.7\%), and 19,211 free joints (8.9\%).
The full category list and per-category counts, as well as more details on dataset construction, human inspection, and physical property estimation, are provided in Appendices~\ref{sec:dataset_details} and~\ref{sec:construction_details}.

\section{Evaluating RealMobility}
\label{sec:evaluating_method}

We first evaluate \method and focus on two questions:
(1) compared with existing single-image articulated generation methods, can \method generate more accurate part structures, geometry, and motion?
(2) compared with agent-based modeling methods, can \method generate assets that are more faithful to the input and have more realistic textures while keeping the motion correct?
All results of \method in this section are generated fully automatically, without any human involvement.

\subsection{Motion Structure Recovery}
\label{sec:compare_single_image}

\paragraph{Experiment details.}
We evaluate on ACD-135~\cite{s2o} using the test split and evaluation protocol of SINGAPO~\cite{singapo}, and reproduce eight single-image articulated generation methods under the same setting as baselines: URDF-Anything+~\cite{urdfanythingplus}, PAct~\cite{pact}, MonoArt~\cite{monoart}, SINGAPO~\cite{singapo}, Articulate-Anything~\cite{articulateanything}, PWM-ArtGen~\cite{pwmartgen}, PhysX-Omni~\cite{physxomni}, and PhysX-Anything~\cite{physxanything}.
All baselines are run with their official code.
For fairness, all methods that use a general-purpose vision-language model, including \method, use GPT-6 Astra~\cite{gpt6astra}.
The metrics are dgIoU and dcDist in the rest state (RS) and articulated states (AS), together with AOR, which measures interpenetration between parts; lower is better for all metrics.

\begin{table}[t]
  \centering
  \caption{Articulated asset generation on ACD-135 under the SINGAPO protocol. Lower is better for all metrics. The best result in each column is in \textbf{bold} and the second best is \underline{underlined}.}
  \label{tab:method_acd}
  \footnotesize
  \setlength{\tabcolsep}{3pt}
  \renewcommand{\arraystretch}{1.05}
  \resizebox{\columnwidth}{!}{%
  \begin{tabular}{@{}lccccc@{}}
    \toprule
    & \multicolumn{4}{c}{Reconstruction quality $\downarrow$} & Collision \\
    \cmidrule(lr){2-5}\cmidrule(l){6-6}
    Method & RS-$d_{\mathrm{gIoU}}$ & AS-$d_{\mathrm{gIoU}}$ & RS-$d_{\mathrm{cDist}}$ & AS-$d_{\mathrm{cDist}}$ & AOR $\downarrow$ \\
    \midrule
    URDF-Anything+ & \underline{0.622} & \underline{0.626} & 0.143 & 0.197 & 0.1940 \\
    PAct & 0.632 & 0.640 & \underline{0.093} & 0.129 & 0.0516 \\
    MonoArt & 0.644 & 0.648 & 0.168 & 0.225 & 0.3667 \\
    SINGAPO & 0.671 & 0.674 & 0.101 & \underline{0.124} & 0.0129 \\
    Articulate-Anything & 0.876 & 0.883 & 0.157 & 0.190 & 0.0118 \\
    PWM-ArtGen & 1.095 & 1.091 & 0.266 & 0.253 & 0.0929 \\
    PhysX-Omni & 1.119 & 1.125 & 0.204 & 0.253 & 0.0082 \\
    PhysX-Anything & 1.255 & 1.256 & 0.238 & 0.279 & \underline{0.0077} \\
    \midrule
    \textbf{\method (ours)} & \textbf{0.517} & \textbf{0.525} & \textbf{0.060} & \textbf{0.072} & \textbf{0.0041} \\
    \bottomrule
  \end{tabular}%
  }
\end{table}

\paragraph{Results.}
As shown in Table~\ref{tab:method_acd}, \method achieves the best results on all five metrics.
On dgIoU, which reflects part structure, \method reaches an RS-dgIoU of 0.517, about 17\% lower than the second-best method.
For geometric accuracy, its RS-dcDist is 0.060, about 35\% lower than the second-best method.
The AS-dcDist grows with errors in joint axes and motion ranges.
\method achieves an AS-dcDist of 0.072, about 42\% lower than the second-best method, and this margin is larger than in the rest state, indicating more accurate joint estimation.
Meanwhile, its AOR is only 0.0041, showing almost no interpenetration between parts.
Qualitative results, analysis, and ablations of the key components of \method are provided in Appendices~\ref{sec:detail_motion} and~\ref{sec:ablation}.

\subsection{Visual Fidelity}
\label{sec:compare_agent}

\paragraph{Experiment details.}
This experiment evaluates whether \method achieves higher fidelity than agent-based modeling methods while keeping a reliable motion structure.
Through a user study, we compare \method with two agent-based modeling methods: GPT-6 Astra + Codex~\cite{codex} and GPT-6 Astra + Articraft~\cite{articraft}.
The study covers 25 everyday categories from real life with five samples per category, and all three methods take the same reference image as input.
We invite 30 participants; each receives one randomly drawn sample from every category, for 25 questions in total.
Participants view the anonymized results of the three methods in an interactive 3D viewer and evaluate them in three aspects.
Motion correctness is judged for each result independently: a result is correct only if the number and types of joints and the positions and directions of the axes are all correct, and the parts are accurately segmented.
Input fidelity and texture quality are three-way choices: for the former, participants select the result most similar to the reference image; for the latter, they ignore the reference image and select the result with the best-looking textures.
More experiment details and results are provided in Appendix~\ref{sec:detail_visual}.

\begin{table}[t]
  \centering
  \caption{User study on 125 objects from 25 categories (\%). Motion correctness is the fraction of assets judged to move correctly; fidelity and texture are the shares of votes each method receives. The best result in each column is in \textbf{bold} and the second best is \underline{underlined}.}
  \label{tab:user_study}
  \footnotesize
  \setlength{\tabcolsep}{1.5pt}
  \renewcommand{\arraystretch}{1.05}
  \begin{tabular*}{\columnwidth}{@{\extracolsep{\fill}}lccc@{}}
    \toprule
    & Motion & \multicolumn{2}{c}{Visual preference $\uparrow$} \\
    \cmidrule(lr){2-2}\cmidrule(l){3-4}
    Method & Correct $\uparrow$ & Fidelity & Texture \\
    \midrule
    GPT-6 Astra + Codex & \underline{87.7} & 4.8 & 4.8 \\
    GPT-6 Astra + Articraft & \textbf{88.8} & \underline{18.3} & \underline{15.5} \\
    \midrule
    \textbf{\method (ours)} & 86.1 & \textbf{76.9} & \textbf{79.7} \\
    \bottomrule
  \end{tabular*}
\end{table}

\begin{figure}[t]
  \centering
  \includegraphics[width=\linewidth]{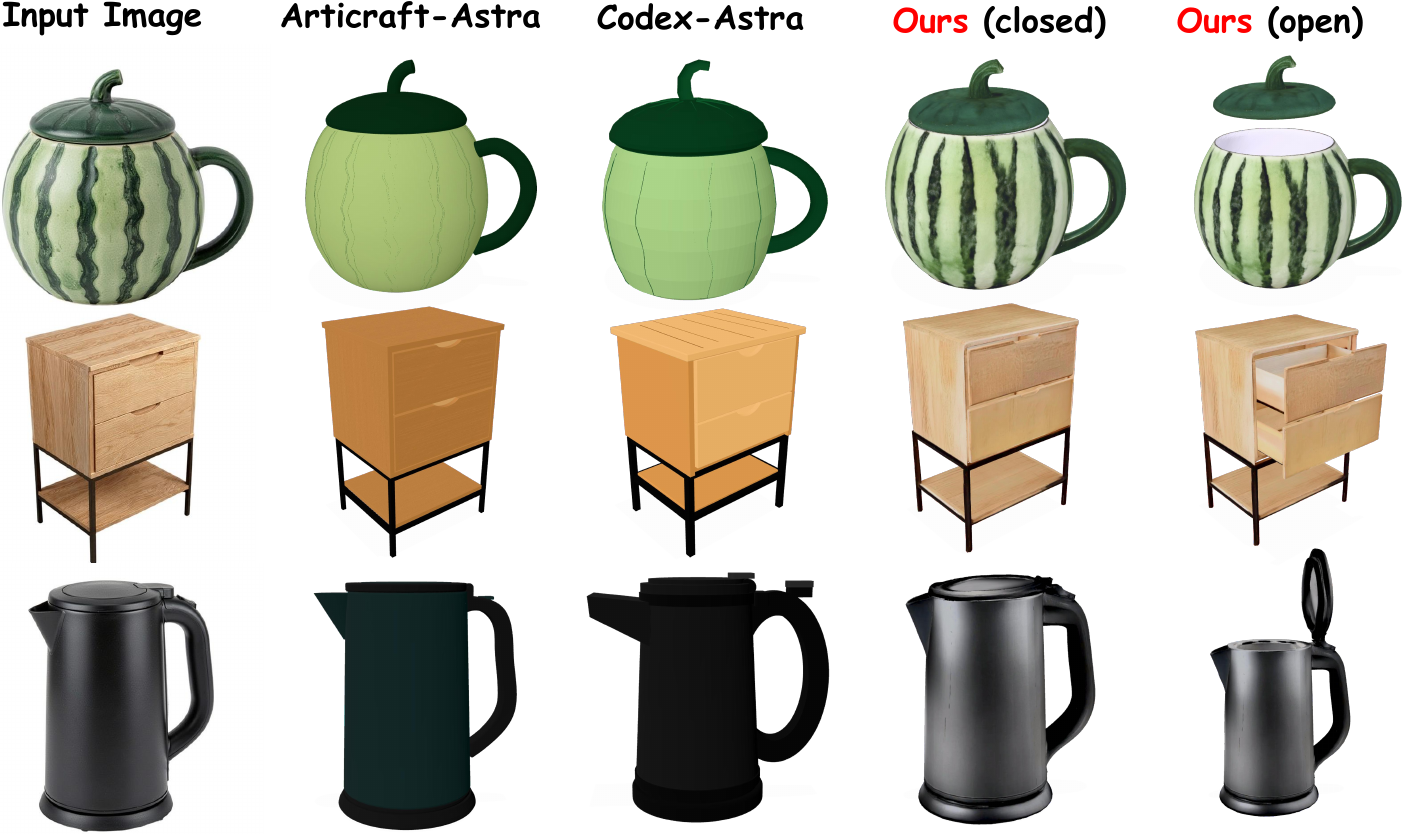}
  \caption{Qualitative comparison with agent-based modeling on user-study samples. Agent-based methods produce simplified geometry with plain colors, whereas \method preserves the shape and texture of the input image.}
  \label{fig:user_study}
\end{figure}

\paragraph{Results.}
As shown in Table~\ref{tab:user_study}, \method reaches 86.1\% motion correctness, within 3 percentage points of the agent-based methods at 88.8\% and 87.7\%.
For input fidelity and texture quality, \method receives 76.9\% and 79.7\% of the votes, versus 18.3\% and 15.5\% for Articraft and 4.8\% each for Codex.
\method thus comes close to agent-based methods in motion correctness while leading by a large margin in fidelity and texture quality.
Qualitative examples are shown in Fig.~\ref{fig:user_study}.

\section{Evaluating RealMobility-117K}
\label{sec:evaluating_dataset}

\begin{figure}[t]
  \centering
  \includegraphics[width=\linewidth]{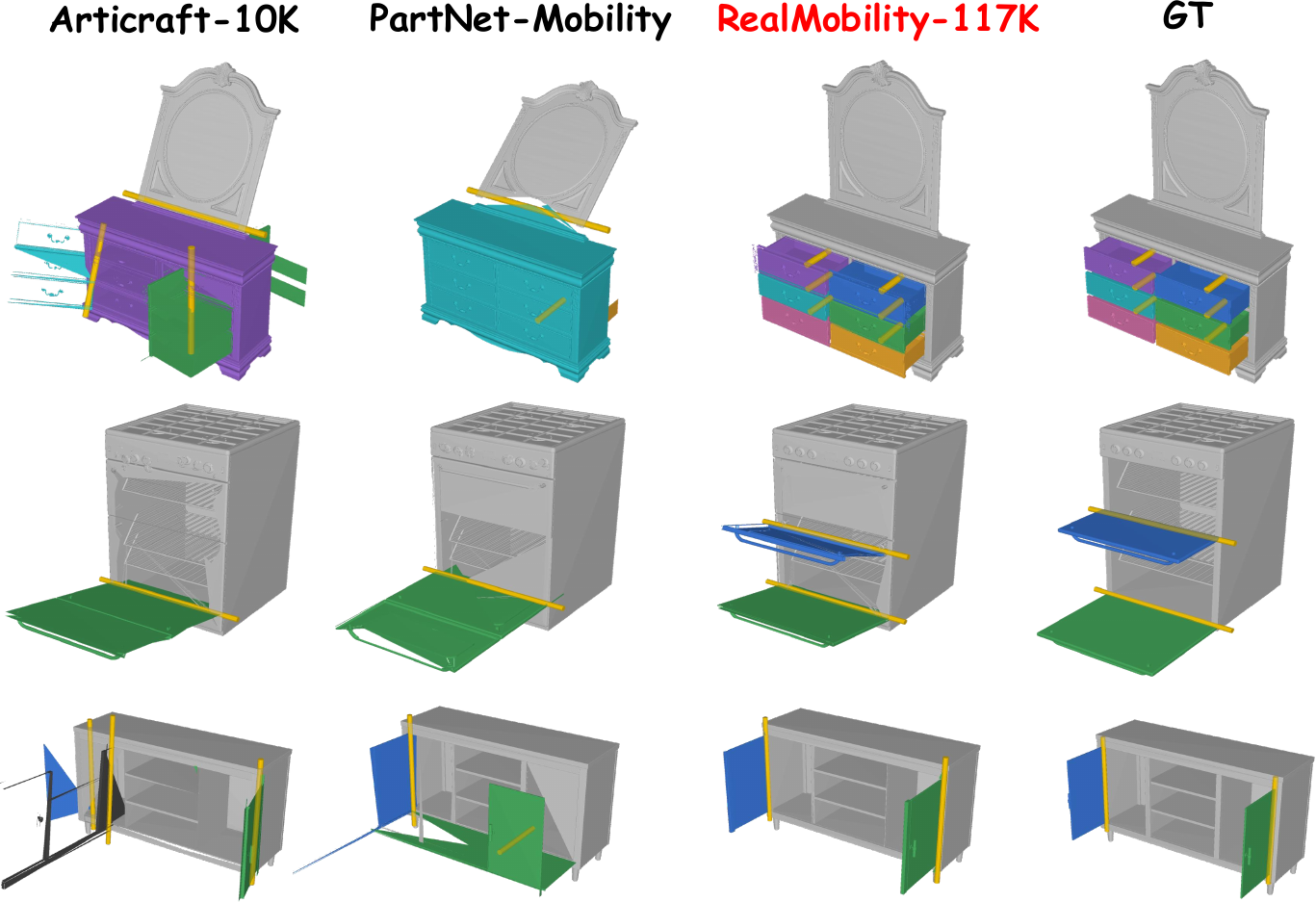}
  \caption{Qualitative results of Particulate on ACD-135 trained with different data. All joints are opened to the upper limit of their predicted ranges. Predicted parts take the color of the ground-truth part they are matched to, and yellow lines mark the predicted rotation axes and sliding directions.}
  \label{fig:qual_acd}
\end{figure}

In this section, we further evaluate \dataset with two questions:
(1) as training data, can \dataset provide more effective supervision for articulation learning than existing datasets?
(2) can its assets and annotations be used for robot interaction in simulation?

\subsection{Articulation Learning}
\label{sec:articulation_learning}

\paragraph{Experiment details.}
We use Particulate~\cite{particulate} as the downstream model to evaluate the training value of \dataset.
We compare three groups of training data: all 2,087 PartNet-Mobility~\cite{sapien} assets in the official training list of Particulate, all 9,854 eligible assets from Articraft-10K~\cite{articraft}, and the same number of 9,854 assets sampled from \dataset.
For sampling, each category receives a quota proportional to its share in the dataset, and assets are drawn randomly within each category.
Only the training data differ across the three groups; the model architecture, hyperparameters, and training steps are identical.
We also test on ACD-135 with the official evaluation code of Particulate.
The metrics are gIoU, mIoU, and PC in the rest state, and gIoU, PC, and OC in the open state.

\begin{table}[t]
  \centering
  \caption{Particulate results on ACD-135 with different training data. Our training set matches Articraft-10K in size. The best result in each column is in \textbf{bold} and the second best is \underline{underlined}.}
  \label{tab:preliminary_learning}
  \footnotesize
  \setlength{\tabcolsep}{1.5pt}
  \renewcommand{\arraystretch}{1.05}
  \begin{tabular*}{\columnwidth}{@{\extracolsep{\fill}}lcccccc@{}}
    \toprule
    & \multicolumn{3}{c}{Rest state} & \multicolumn{3}{c}{Open state} \\
    \cmidrule(lr){2-4}\cmidrule(l){5-7}
    Training data & gIoU $\uparrow$ & PC $\downarrow$ & mIoU $\uparrow$ & gIoU $\uparrow$ & PC $\downarrow$ & OC $\downarrow$ \\
    \midrule
    PartNet-Mobility & \underline{0.295} & \underline{0.109} & \underline{0.464} & \underline{0.282} & \underline{0.166} & \underline{0.0071} \\
    Articraft-10K & 0.171 & 0.130 & 0.390 & 0.164 & 0.199 & 0.0076 \\
    \midrule
    \textbf{\dataset} & \textbf{0.452} & \textbf{0.071} & \textbf{0.578} & \textbf{0.439} & \textbf{0.143} & \textbf{0.0059} \\
    \bottomrule
  \end{tabular*}
\end{table}

\paragraph{Results.}
As shown in Table~\ref{tab:preliminary_learning}, the model trained on \dataset achieves the best results on all six metrics.
For part structure, its rest-state gIoU reaches 0.452, 2.6 times that of Articraft-10K (0.171) and 53\% higher than that of PartNet-Mobility (0.295).
Its mIoU is 0.578, higher than 0.464 for PartNet-Mobility and 0.390 for Articraft-10K, and its rest-state PC is 0.071, 35\% lower than that of PartNet-Mobility (0.109).
The open-state PC and OC reflect whether parts are placed correctly after motion.
Training on \dataset reduces the open-state PC to 0.143, 14\% and 28\% lower than PartNet-Mobility and Articraft-10K, and the OC to 0.0059, 17\% and 22\% lower, respectively.
This indicates that the model learns not only more accurate part segmentation but also more accurate part motion.
Notably, among the 16 categories of ACD-135, the model trained on \dataset achieves the highest rest-state and open-state gIoU in 14 of them.
This shows that the gain does not come from a few categories but holds for most everyday objects, reflecting the broad category and structure coverage of the dataset; full per-category metrics are provided in Appendix~\ref{sec:detail_learning}.
Qualitative results are shown in Fig.~\ref{fig:qual_acd}.
These results show that \dataset provides more effective supervision for articulation learning than existing datasets.

\subsection{Robot Simulation}
\label{sec:robot_simulation}

\begin{figure}[t]
  \centering
  \includegraphics[width=\linewidth]{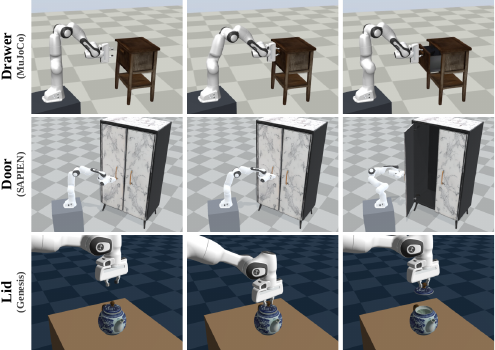}
  \caption{Robot interaction with \dataset assets in three simulators.}
  \label{fig:sim}
\end{figure}

We export assets to URDF and MJCF with our conversion tools and load them into MuJoCo, SAPIEN, and Genesis.
For each asset, we plan a target trajectory for one movable part and use inverse kinematics to drive a Franka arm that grasps the handle and operates the part, holding it by friction alone.
As shown in Fig.~\ref{fig:sim}, the arm pulls out a drawer in MuJoCo, opens a wardrobe door in SAPIEN, and lifts a teapot lid in Genesis, and all three types of movable parts are operated successfully.
These examples demonstrate the usability of \dataset in robot simulation.

\section{Conclusion}
\label{sec:conclusion}

We present \method, a training-free framework that uses a pixel-aligned 3D generator to turn motion in generated videos into part and joint structures while keeping detailed geometry and textures.
Using it, we build \dataset, which contains 117,130 human-verified articulated assets with structural, semantic, and physical property annotations.
Experiments show that \method outperforms single-image articulated generation methods in motion structure recovery and agent-based methods in visual quality, and that \dataset provides more effective supervision for articulation learning than existing datasets and is usable in robot simulation.

\section*{Acknowledgments}
This work is supported by the Shenzhen Loop Area Institute under grant FPF10120260001.

{
    \small
    \bibliographystyle{ieeenat_fullname}
    \bibliography{main}
}

\clearpage
\appendix
\setcounter{table}{0}
\setcounter{figure}{0}
\renewcommand{\thetable}{S\arabic{table}}
\renewcommand{\thefigure}{S\arabic{figure}}
\phantomsection
\section*{Appendix}
\label{sec:overview}

This appendix provides additional experiments and more details on the main text.
All experiments use \ours{} and \data{} exactly as in the main text; no component is re-tuned.
Table~\ref{tab:index} lists where to find the content referenced in the main text.

\begin{table}[!htbp]
\centering
\small
\caption{Where to find the content referenced in the main text (Main) and additional content in this appendix (App.).}
\label{tab:index}
\begin{tabular*}{\linewidth}{@{\extracolsep{\fill}}lcc@{}}
\toprule
Content & Main & App. \\
\midrule
Choice of the axis-fitting method & Sec.~3.2 & \ref{sec:joint_fitting} \\
Interior completion of drawers & Sec.~3.3 & \ref{sec:interior} \\
Inspection and physical properties & Sec.~4 & \ref{sec:construction_details} \\
Full category list and counts & Sec.~4 & \ref{sec:dataset_details} \\
Ablation study & Sec.~5.1 & \ref{sec:ablation} \\
Motion structure recovery details & Sec.~5.1 & \ref{sec:detail_motion} \\
User study details & Sec.~5.2 & \ref{sec:detail_visual} \\
Per-category learning results & Sec.~6.1 & \ref{sec:detail_learning} \\
Comparison with FreeArt3D & -- & \ref{sec:cmp_prior} \\
More dataset examples & -- & \ref{sec:examples} \\
\bottomrule
\end{tabular*}
\end{table}

\section{More Details on RealMobility}
\label{sec:method_details}

This section gives more details on five parts of \ours{} (Sec.~\ref{sec:method}):
keyframe selection (Sec.~\ref{sec:keyframe}),
cross-state voting and part segmentation (Sec.~\ref{sec:voting}),
axis fitting for revolute joints (Sec.~\ref{sec:joint_fitting}),
the choice of the reference state (Sec.~\ref{sec:ref_state}),
and interior completion of drawers (Sec.~\ref{sec:interior}).
All distances in voxel space are measured in voxels of the $64^3$ grid of CUPID~\cite{cupid}.

\subsection{Keyframe Selection}
\label{sec:keyframe}

In a generated video, a part may barely move in the first few seconds and then open quickly.
If frames are sampled uniformly in time, most keyframes fall in the slow stages, where adjacent states are almost the same, while the fast stage is skipped.

To select keyframes by motion progress (Sec.~\ref{sec:motion_synthesis}), we use CoWTracker~\cite{cowtracker} to track pixels through the $N$ frames of the video.
Pixels with large displacements form the moving region, and the other reliably tracked pixels form the static region.
Let $d_i$ be the median displacement of the moving region from frame $i-1$ to frame $i$ minus the median displacement of the static region;
negative values are set to zero, and $d_i$ is smoothed over five frames.
The motion progress of frame $i$ is
\begin{equation}
  p_i = \frac{\sum_{j=1}^{i} d_j}{\sum_{j=1}^{N-1} d_j}, \qquad p_0 = 0.
  \label{eq:progress}
\end{equation}
We take six evenly spaced progress values $k/5$, $k = 0,\dots,5$, and use the first frame that reaches each value as the $k$-th keyframe:
\begin{equation}
  i_k = \min \{\, i : p_i \geq k/5 \,\}.
  \label{eq:keyframe}
\end{equation}
Each keyframe may also move by up to two frames, to a frame with higher tracking confidence at almost the same progress.
For videos of closing motions, the order of the keyframes is reversed so that state 0 is the closed state.
The resulting keyframes are evenly distributed over the motion, and adjacent states differ by clear motion.
This gives stable inputs for cross-state voting and axis fitting.

\subsection{Cross-state Voting and Part Segmentation}
\label{sec:voting}

Cross-state voting (Eq.~\eqref{eq:state_support}) assumes that the base stays the same across states.

However, the interior of an object can change a lot with its motion state, which disturbs the voting.
Take a cabinet as an example: when the door is closed, CUPID often reconstructs the inside of the cabinet as solid voxels; after the door opens, the cavity is exposed and the reconstruction becomes hollow.
These interior voxels appear only in some states and thus receive few votes.
They are then mistaken for movable-part candidates and contaminate the segmented movable parts.

\paragraph{States, alignment, and voting.}
In the experiments, the parts are open in the input image, and the generated video closes them, so the keyframes are reversed (Sec.~\ref{sec:keyframe}).
For the alignment (Eq.~\eqref{eq:state_alignment}), the UV coordinates are sampled on a $20 \times 20$ image grid.
A state is discarded if its number of inliers is below 0.7 times the average over the states.

\paragraph{Interior filtering.}
We remove interior structures at two points.
Before clustering, we estimate the front of the base from the bounding box of the closed state, and remove the candidates in the open states that lie more than 2 voxels behind this front:
opened doors and pulled-out drawers lie in front of the base, while interior structures lie behind it.
After clustering, a cluster is regarded as interior and skipped if its centroid, projected along the vertical axis, lies within 1.5 voxels of the footprint of the closed state.
A drawer cluster must also stick out from the base by at least 2 voxels.

\paragraph{Clustering.}
In each state, the remaining candidates are clustered with DBSCAN (radius 1.5, at least 10 neighbors).
A point is also a core point if at least 8 of its neighbors lie on one plane with a thickness of at most 0.3,
so that door panels only one voxel thick are not lost.

\subsection{Axis Fitting for Revolute Joints}
\label{sec:joint_fitting}

\ours{} obtains the axis of a revolute joint in one of two ways.
Both solve for the axis direction $a$ and a point $c$ on the axis in Eq.~\eqref{eq:shared_axis}.
Since CUPID reconstructs objects upright and the axes of furniture and appliances are parallel to the coordinate axes, in practice we choose $a$ only among the three coordinate axes.

\paragraph{Tracking-based fitting.}
This approach works for parts of any shape but depends on the tracking quality: tracks become unreliable when the part has little texture or is occluded during the motion.
CoWTracker tracks the keyframe of each valid state to every other valid state.
Pixels with visibility above 0.1 are mapped onto the segmented part through the UV maps, which gives the positions $(x^p_t, x^p_{t'})$ of the same part point in two states.
Correspondences that move less than 5 voxels are removed, since they carry little information about the rotation.
If the axis is correct, every correspondence satisfies Eq.~\eqref{eq:shared_axis}: rotating $x^p_t$ about the axis $(a, c)$ by the angle $\theta_{tt'}$ brings it to $x^p_{t'}$.
For each candidate direction $a$, we first estimate $c$ and $\theta_{tt'}$ for each pair of states with RANSAC and remove the correspondences with errors above 1 voxel.
We then solve Eq.~\eqref{eq:shared_axis} jointly on the remaining correspondences.

\paragraph{Plane intersection.}
A flat part such as a door panel is approximately a plane in each state.
These planes rotate about the same axis, so their common intersection line is the axis.
This approach does not rely on tracking and only requires fitting planes.
The closed state, in which the part touches the base, is not used.
Since the axis lies in the plane of the panel, the panel becomes a line in every state when viewed along the axis.
Therefore, for each coordinate axis, we project the part points of each state along this axis onto 2D and fit a line for each state;
the coordinate axis with the smallest line-fitting error is taken as $a$.
In this 2D plane, the lines of different states should meet at one point, which is the position $c$ of the axis.
In practice, the intersection points do not coincide exactly.
We choose the state whose intersections with the other states are the most concentrated, and take the mean of its intersection points as $c$.

\paragraph{Automatic choice.}
Plane intersection relies on two conditions: the part is a flat panel, and the axis lies in the panel, so that the orientation of the panel changes during the rotation.
Otherwise, it cannot give a reliable axis.
For example, the lid of a swivel lighter rotates within its own plane, about an axis perpendicular to the lid.
Ideally, the lid planes of all states are the same plane, so no intersection line can be found.
For thick or curved parts, the fitted planes themselves are unreliable.
Therefore, for each revolute joint, we use plane intersection only if the part is a flat panel and its plane does rotate:
the median ratio of the smallest to the second-largest PCA eigenvalue of the part points over the valid states is below 0.15,
and the panel normal turns by at least $15^\circ$ between some pair of states.
Otherwise, we use tracking-based fitting.
In the ACD-135 experiments, 165 revolute joints are processed: 118 (72\%) use plane intersection, and 47 (28\%) are passed to tracking-based fitting.

\subsection{Choice of the Reference State}
\label{sec:ref_state}

\ours{} accepts input images with parts either closed or open.

Note that the reference state here differs from the alignment target in Sec.~\ref{sec:alignment_segmentation}.
Within a stage, all states are aligned to state 0, whereas the reference state is the state used to generate the final asset: its voxels are labeled by part and decoded into textured meshes.

By default, we use the most open state, i.e., state 5 of the last stage, as the reference state instead of the closed state, for two reasons.

\emph{Parts are hard to segment in the closed state.}
When closed, a movable part touches the base, and their voxels stick together.
The voxel resolution of CUPID is $64 \times 64 \times 64$, and thin parts such as cabinet doors are often only one voxel thick;
in the closed state, their voxels overlap with or directly neighbor those of the front of the base and cannot be cleanly separated.
Once opened, the movable part leaves the base, so the two are spatially separated and the segmentation is more accurate.

\emph{The most open state provides the most reference for the interior.}
It exposes the most interior structures, such as the inner walls of cabinets and the cavities of drawers.
Since the second step of CUPID generates geometry and textures conditioned on the keyframe of the reference state, choosing this state provides the most reference for generating interior structures.

\subsection{Interior Completion of Drawers}
\label{sec:interior}

\begin{figure}[!htbp]
\centering
\includegraphics[width=\columnwidth]{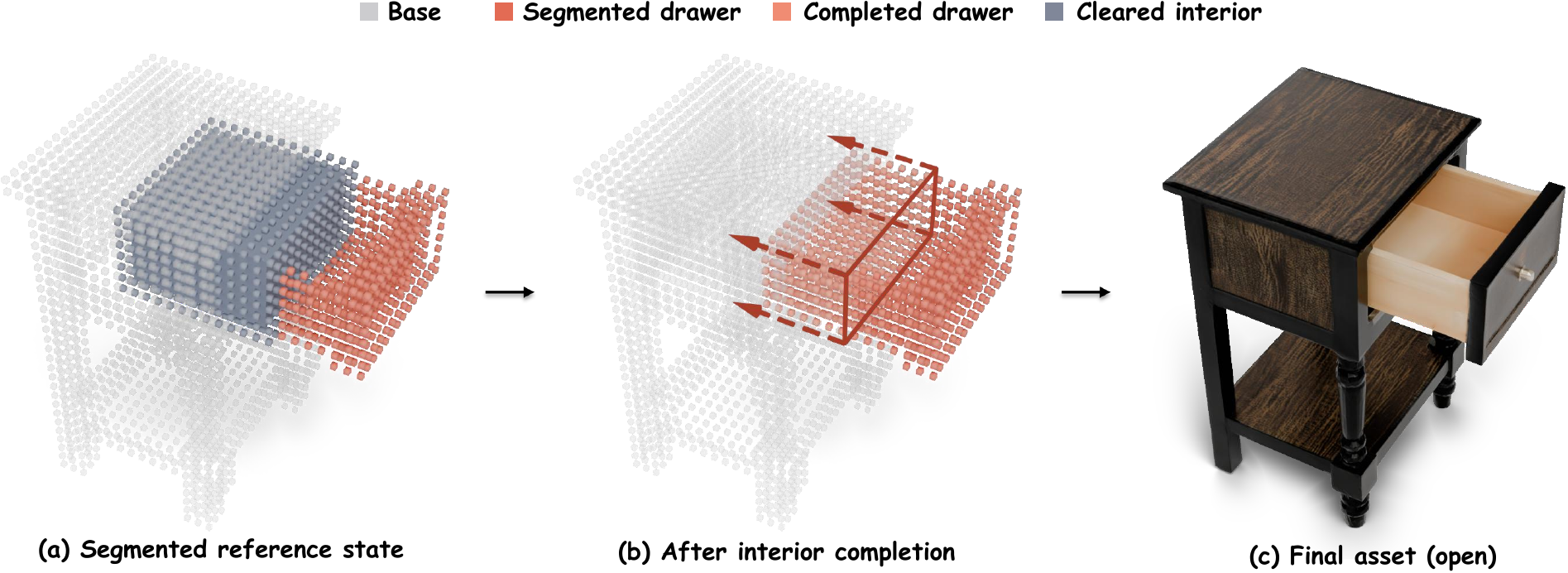}
\caption{Example of interior completion of drawers.}
\label{fig:interior}
\end{figure}

In the generated open state, only part of the drawer front is visible, and the inside of the base is often reconstructed as solid.
If used directly, a drawer is only the part that sticks out, and the base has no space to hold it.
We therefore complete the drawer box on the voxels of the reference state and carve a matching cavity in the base (Sec.~\ref{sec:completion_generation}).
Fig.~\ref{fig:interior} shows an example.

\paragraph{Closed position and depth.}
Both are measured in the aligned closed state 0.
We map each column covered by the drawer front to state 0 and take the outermost base voxel along the sliding direction;
the mode over all columns gives the position of the drawer front when it is closed.
We use the mode so that the handle does not bias the result.
The depth of the base is measured in a similar way, as the length of the base of state 0 along the sliding direction.
If the local depth within the cross-section of a drawer is less than 0.8 times this length, we use the local depth instead, e.g., for a small cabinet on a desk.

\paragraph{Completion.}
The drawer box extends inward from the front by 0.95 times the depth and consists of a bottom, two side panels, and a back panel.
In the base, we carve a cavity of the same depth along the cross-section of the drawer and add the base walls around the cavity.
The base voxels swept by the drawer when it opens are also removed.
Multiple drawers are processed one by one, each with its own closed position.

\paragraph{Motion range.}
In the closed position, the drawer front is flush with the front of the base.
In the fully open position, the drawer is pulled out from the closed position by 0.65 times the box length.
The pull-out distance of the drawer in the generated frames is not used.

\section{More Details on RealMobility-117K}
\label{sec:dataset_details}

\data{} contains 117,130 assets from 89 categories, with 216,508 joints in total: 72,290 revolute joints (33.4\%), 125,007 prismatic joints (57.7\%), and 19,211 free joints (8.9\%).
We group the 89 categories into eight super-categories (Table~\ref{tab:supercat}), and Table~\ref{tab:categories} lists all categories and their numbers of assets.

\begin{table}[!htbp]
\centering
\small
\caption{Super-categories of \data{}.}
\label{tab:supercat}
\begin{tabular}{@{}lrrr@{}}
\toprule
Super-category & \#Categories & \#Assets & Share (\%) \\
\midrule
Cabinets \& storage & 21 & 44,373 & 37.9 \\
Large appliances & 10 & 16,437 & 14.0 \\
Containers \& bins & 11 & 15,282 & 13.0 \\
Kitchen appliances & 13 & 12,117 & 10.3 \\
Tables, beds \& doors & 8 & 10,944 & 9.3 \\
Cookware \& drinkware & 9 & 7,579 & 6.5 \\
Boxes \& cases & 10 & 5,216 & 4.5 \\
Electronics \& gadgets & 7 & 5,182 & 4.4 \\
\midrule
Total & 89 & 117,130 & 100.0 \\
\bottomrule
\end{tabular}
\end{table}

The number of assets per category ranges from 61 (ice maker) to 4,612 (chest of drawers), with a median of 868.
Forty-two categories have more than 1,000 assets, and only three have fewer than 100.
The ten largest categories account for only 29.8\% of all assets, so the data are not concentrated in a few categories.

More visualizations of the assets are shown in Sec.~\ref{sec:examples}.

\begin{table*}[t]
\centering
\small
\caption{All 89 categories of \data{} and their numbers of assets (117,130 in total). Super-categories are in bold, followed by their total numbers of assets.}
\label{tab:categories}
\begin{tabular*}{\textwidth}{@{\extracolsep{\fill}}lrlrlr@{}}
\toprule
Category & \#Assets & Category & \#Assets & Category & \#Assets \\
\midrule
\textbf{Cabinets \& storage} & \textbf{44,373} & \textbf{Containers \& bins} & \textbf{15,282} & \hspace{0.8em}kids' desk & 580 \\
\hspace{0.8em}chest of drawers & 4,612 & \hspace{0.8em}tea canister & 3,052 & \hspace{0.8em}double entry door & 570 \\
\hspace{0.8em}nightstand & 3,889 & \hspace{0.8em}trash can & 2,517 & \textbf{Cookware \& drinkware} & \textbf{7,579} \\
\hspace{0.8em}bathroom vanity & 3,698 & \hspace{0.8em}screw-cap jar & 2,308 & \hspace{0.8em}soup pot & 3,110 \\
\hspace{0.8em}dresser & 3,292 & \hspace{0.8em}wheelie bin & 1,859 & \hspace{0.8em}teapot & 1,520 \\
\hspace{0.8em}single-door cabinet & 3,250 & \hspace{0.8em}ashtray & 1,548 & \hspace{0.8em}insulated tumbler & 1,424 \\
\hspace{0.8em}multi-drawer cabinet & 2,852 & \hspace{0.8em}mailbox & 1,212 & \hspace{0.8em}mug & 463 \\
\hspace{0.8em}filing cabinet & 2,728 & \hspace{0.8em}stoppered bottle & 777 & \hspace{0.8em}baby bottle & 405 \\
\hspace{0.8em}bookcase & 2,691 & \hspace{0.8em}fermentation crock & 659 & \hspace{0.8em}food cloche & 337 \\
\hspace{0.8em}bedroom storage cabinet & 2,649 & \hspace{0.8em}beverage bottle & 621 & \hspace{0.8em}wok & 136 \\
\hspace{0.8em}safe & 2,531 & \hspace{0.8em}file box & 377 & \hspace{0.8em}thermal jug & 114 \\
\hspace{0.8em}wardrobe & 2,174 & \hspace{0.8em}storage bin & 352 & \hspace{0.8em}sports bottle & 70 \\
\hspace{0.8em}TV stand & 1,802 & \textbf{Kitchen appliances} & \textbf{12,117} & \textbf{Boxes \& cases} & \textbf{5,216} \\
\hspace{0.8em}rolling drawer cart & 1,614 & \hspace{0.8em}oven & 3,436 & \hspace{0.8em}watch box & 820 \\
\hspace{0.8em}storage ottoman & 1,312 & \hspace{0.8em}microwave oven & 3,277 & \hspace{0.8em}small wooden box & 637 \\
\hspace{0.8em}fire equipment cabinet & 1,272 & \hspace{0.8em}rice cooker & 2,192 & \hspace{0.8em}shoebox & 619 \\
\hspace{0.8em}display cabinet & 1,006 & \hspace{0.8em}air fryer & 1,146 & \hspace{0.8em}soap dish box & 566 \\
\hspace{0.8em}kitchen base cabinet & 981 & \hspace{0.8em}electric kettle & 520 & \hspace{0.8em}sewing box & 553 \\
\hspace{0.8em}kitchen wall cabinet & 973 & \hspace{0.8em}sandwich maker & 323 & \hspace{0.8em}jewelry box & 501 \\
\hspace{0.8em}mirror cabinet & 640 & \hspace{0.8em}food dehydrator & 301 & \hspace{0.8em}powder compact & 416 \\
\hspace{0.8em}shoe cabinet & 229 & \hspace{0.8em}juicer & 227 & \hspace{0.8em}vintage trunk & 378 \\
\hspace{0.8em}wine cabinet & 178 & \hspace{0.8em}blender & 208 & \hspace{0.8em}cigar box & 373 \\
\textbf{Large appliances} & \textbf{16,437} & \hspace{0.8em}soymilk maker & 198 & \hspace{0.8em}glasses case & 353 \\
\hspace{0.8em}top-load washer & 3,321 & \hspace{0.8em}waffle maker & 152 & \textbf{Electronics \& gadgets} & \textbf{5,182} \\
\hspace{0.8em}front-load washer & 2,808 & \hspace{0.8em}clamshell grill & 76 & \hspace{0.8em}lighter & 1,516 \\
\hspace{0.8em}dehumidifier & 1,953 & \hspace{0.8em}ice maker & 61 & \hspace{0.8em}laptop & 1,314 \\
\hspace{0.8em}refrigerator & 1,801 & \textbf{Tables, beds \& doors} & \textbf{10,944} & \hspace{0.8em}paper shredder & 868 \\
\hspace{0.8em}dish sterilizer & 1,738 & \hspace{0.8em}interior door & 2,719 & \hspace{0.8em}printer & 715 \\
\hspace{0.8em}chest freezer & 1,695 & \hspace{0.8em}desk with drawers & 2,383 & \hspace{0.8em}record player & 339 \\
\hspace{0.8em}clothes dryer & 1,112 & \hspace{0.8em}coffee table & 1,661 & \hspace{0.8em}foldable phone & 218 \\
\hspace{0.8em}dishwasher & 804 & \hspace{0.8em}dressing table & 1,545 & \hspace{0.8em}flip phone & 212 \\
\hspace{0.8em}pet dryer box & 784 & \hspace{0.8em}side table & 792 &  &  \\
\hspace{0.8em}mini washer & 421 & \hspace{0.8em}storage bed & 694 &  &  \\
\bottomrule
\end{tabular*}
\end{table*}

\section{More Details on Experiments}
\label{sec:exp_details}

This section provides more details on three experiments of the main text:
motion structure recovery (Sec.~\ref{sec:detail_motion}),
visual fidelity (Sec.~\ref{sec:detail_visual}),
and articulation learning (Sec.~\ref{sec:detail_learning}).

\subsection{Motion Structure Recovery}
\label{sec:detail_motion}

\paragraph{Metrics.}
We use the evaluation code of SINGAPO~\cite{singapo}.
Each part is represented by its bounding box.
$d_\text{gIoU}$ is defined as $1 - \text{gIoU}$ and measures the difference between the bounding boxes of the predicted and ground-truth parts; $d_\text{cDist}$ measures the distance between their centers.
Predicted and ground-truth parts are matched one-to-one with the Hungarian algorithm, and each metric is computed both from the prediction to the ground truth and from the ground truth to the prediction, and then averaged.
RS reflects the static part segmentation and geometry.
AS moves each part with its own joint and averages over five evenly spaced states, so it also reflects errors in the joint axes and motion ranges.
AOR takes ten states of the same motion, computes the overlap ratio between the bounding boxes of sibling parts under the same parent, and averages them to measure interpenetration between parts.

\paragraph{Input state.}
\ours{} accepts an input image with parts either open or closed, but in both cases the open state is used as the reference when the second step of CUPID generates the final result (Sec.~\ref{sec:ref_state}).
To compare with a closed-state ground truth, the parts would first have to be moved back to the closed pose with the predicted joint axes, which would add axis errors to RS.
RS is a static segmentation metric that does not involve the joint axes, and axis errors are measured by AS.
We therefore use open-state inputs for \ours{}, so that its output and the ground truth are in the same pose and RS evaluates only the static segmentation.
For the other methods, we test both open-state and closed-state inputs, and Table~\ref{tab:method_acd} reports the better of the two inputs for each metric separately.
Table~\ref{tab:input_state} lists the results of each method with both inputs.

\begin{table*}[t]
\centering
\small
\caption{Results on ACD-135 with open-state and closed-state inputs under the SINGAPO protocol. Lower is better for all metrics. For each baseline, Table~\ref{tab:method_acd} reports the better of the two inputs for each metric separately (underlined). MonoArt and URDF-Anything+ output meshes in the observed pose with zero joint values. With open-state inputs, it is unknown how far each part is opened, so AS and AOR are undefined (--).}
\label{tab:input_state}
\begin{tabular}{@{}llccccc@{}}
\toprule
Method & Input & RS-$d_\text{gIoU}$ & AS-$d_\text{gIoU}$ & RS-$d_\text{cDist}$ & AS-$d_\text{cDist}$ & AOR \\
\midrule
\multirow{2}{*}{URDF-Anything+~\cite{urdfanythingplus}} & Open & 0.775 & -- & 0.211 & -- & -- \\
 & Closed & \underline{0.622} & \underline{0.626} & \underline{0.143} & \underline{0.197} & \underline{0.1940} \\
\cmidrule{1-7}
\multirow{2}{*}{PAct~\cite{pact}} & Open & 1.007 & 0.998 & 0.180 & 0.161 & 0.0656 \\
 & Closed & \underline{0.632} & \underline{0.640} & \underline{0.093} & \underline{0.129} & \underline{0.0516} \\
\cmidrule{1-7}
\multirow{2}{*}{MonoArt~\cite{monoart}} & Open & 0.856 & -- & 0.250 & -- & -- \\
 & Closed & \underline{0.644} & \underline{0.648} & \underline{0.168} & \underline{0.225} & \underline{0.3667} \\
\cmidrule{1-7}
\multirow{2}{*}{SINGAPO~\cite{singapo}} & Open & 0.912 & 0.906 & 0.155 & 0.138 & 0.0257 \\
 & Closed & \underline{0.671} & \underline{0.674} & \underline{0.101} & \underline{0.124} & \underline{0.0129} \\
\cmidrule{1-7}
\multirow{2}{*}{Articulate-Anything~\cite{articulateanything}} & Open & 1.048 & 1.037 & 0.210 & \underline{0.190} & \underline{0.0118} \\
 & Closed & \underline{0.876} & \underline{0.883} & \underline{0.157} & 0.202 & 0.0171 \\
\cmidrule{1-7}
\multirow{2}{*}{PWM-ArtGen~\cite{pwmartgen}} & Open & \underline{1.095} & \underline{1.091} & \underline{0.266} & \underline{0.253} & \underline{0.0929} \\
 & Closed & 1.127 & 1.132 & 0.336 & 0.354 & 0.1348 \\
\cmidrule{1-7}
\multirow{2}{*}{PhysX-Omni~\cite{physxomni}} & Open & 1.313 & 1.308 & 0.343 & 0.298 & \underline{0.0082} \\
 & Closed & \underline{1.119} & \underline{1.125} & \underline{0.204} & \underline{0.253} & 0.0140 \\
\cmidrule{1-7}
\multirow{2}{*}{PhysX-Anything~\cite{physxanything}} & Open & 1.341 & 1.339 & 0.365 & 0.320 & 0.0079 \\
 & Closed & \underline{1.255} & \underline{1.256} & \underline{0.238} & \underline{0.279} & \underline{0.0077} \\
\cmidrule{1-7}
\ours{} (ours) & Open & 0.517 & 0.525 & 0.060 & 0.072 & 0.0041 \\
\bottomrule
\end{tabular}
\end{table*}

\paragraph{Failed samples.}
If a method does not output a valid articulated object for a sample, the sample is counted as a failure and assigned the worst values, \ie, $d_\text{gIoU} = 2$ and $d_\text{cDist} = 1$, so the denominator is always all 135 samples; AOR is averaged over the successful samples only.
\ours{} succeeds on all 135 samples.
Table~\ref{tab:success_only} reports the results computed only on the successful samples of each method.
Even without counting the failures, \ours{} remains the best on all metrics.

\begin{table}[!htbp]
\centering
\footnotesize
\setlength{\tabcolsep}{3pt}
\caption{Results on ACD-135 computed only on the successful samples of each method. Lower is better for all metrics. For each baseline and each metric, the better of the two inputs on the successful samples is taken. Succ.: number of successful samples with closed / open inputs (out of 135). AOR in Table~\ref{tab:method_acd} is already averaged over the successful samples and is omitted here.}
\label{tab:success_only}
\begin{tabular*}{\linewidth}{@{\extracolsep{\fill}}lccccc@{}}
\toprule
Method & Succ. & RS-$d_\text{gIoU}$ & AS-$d_\text{gIoU}$ & RS-$d_\text{cDist}$ & AS-$d_\text{cDist}$ \\
\midrule
URDF-Anything+ & 135 / 135 & 0.622 & 0.626 & 0.143 & 0.197 \\
PAct & 135 / 133 & 0.632 & 0.640 & 0.093 & 0.129 \\
MonoArt & 135 / 135 & 0.644 & 0.648 & 0.168 & 0.225 \\
SINGAPO & 135 / 135 & 0.671 & 0.674 & 0.101 & 0.124 \\
Articulate-Anything & 134 / 134 & 0.868 & 0.875 & 0.151 & 0.184 \\
PWM-ArtGen & 104 / 122 & 0.867 & 0.873 & 0.138 & 0.162 \\
PhysX-Omni & 132 / 127 & 1.099 & 1.105 & 0.186 & 0.236 \\
PhysX-Anything & 129 / 128 & 1.221 & 1.222 & 0.203 & 0.246 \\
\midrule
\ours{} (ours) & -- / 135 & \textbf{0.517} & \textbf{0.525} & \textbf{0.060} & \textbf{0.072} \\
\bottomrule
\end{tabular*}
\end{table}

\begin{figure*}[!t]
\centering
\includegraphics[width=\textwidth,height=0.9\textheight,keepaspectratio,trim=0 0.45in 0 0,clip]{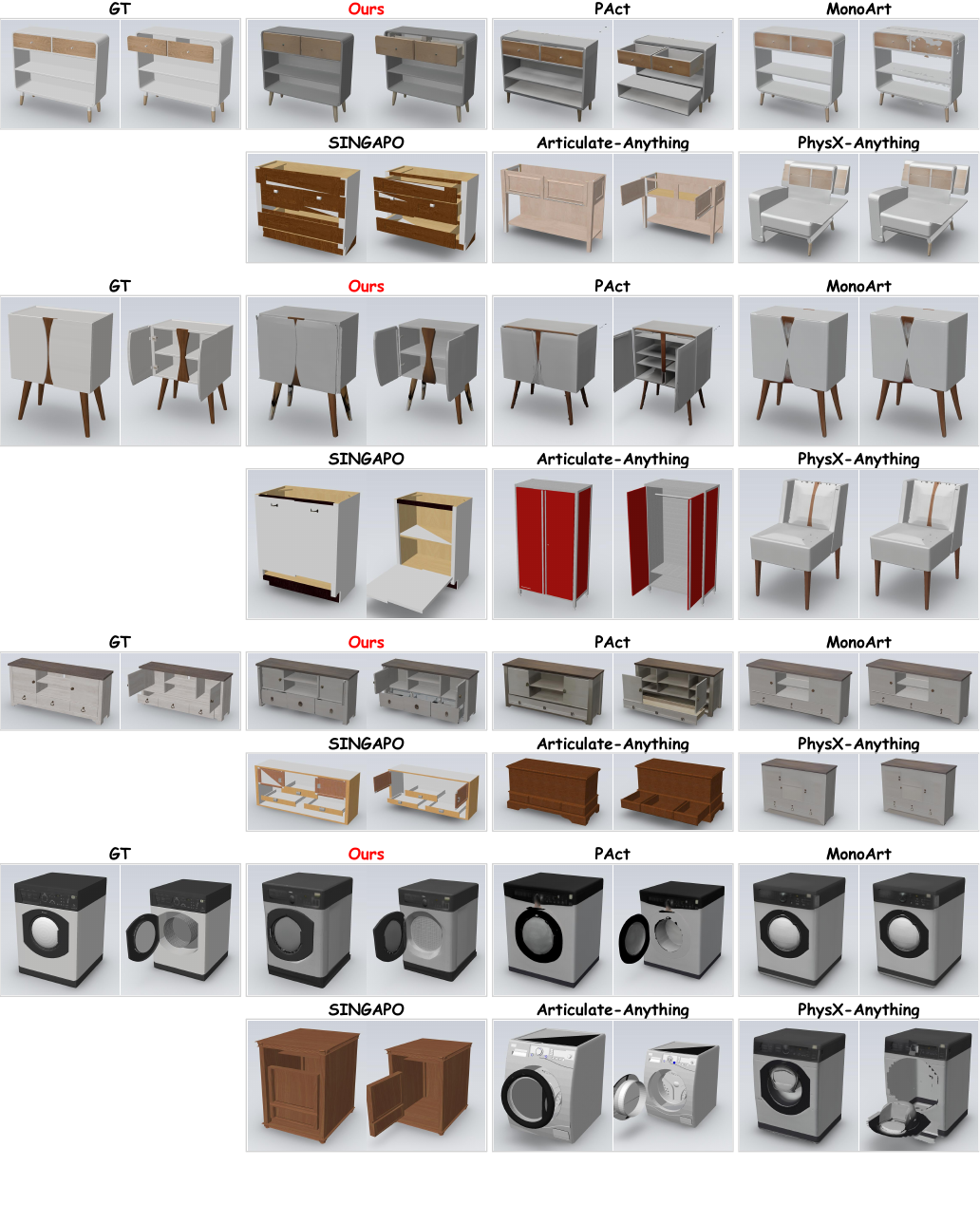}
\caption{More qualitative comparisons on ACD-135.
Each object takes two rows: the first row shows the ground truth, \ours{}, PAct, and MonoArt, and the second row shows SINGAPO, Articulate-Anything, and PhysX-Anything.
For each method, the left and right images show the rest state and the articulated state, respectively.
\ours{} keeps the shape and texture of the input and moves all movable parts, while the baselines often produce mismatched shapes, wrong categories, or parts that do not move.}
\label{fig:qual_acd_more}
\end{figure*}

\paragraph{More qualitative results and analysis.}
Fig.~\ref{fig:qual_acd_more} shows more qualitative comparisons on ACD-135.
The retrieval-based methods, SINGAPO and Articulate-Anything, often produce shapes that do not match the input; the parts of MonoArt barely move in the articulated state; PhysX-Anything sometimes generates objects of the wrong category; and PAct may produce extra movable parts.
In contrast, \ours{} produces geometry and textures consistent with the input and recovers the motion of all movable parts.

\subsection{Visual Fidelity}
\label{sec:detail_visual}

\paragraph{Samples.}
We choose 25 common categories and, for each category, collect five images of objects with different structures from the Internet.
Web images often contain cluttered backgrounds, other objects, and items inside containers such as cabinets and drawers.
We therefore regenerate each image with FLUX~\cite{flux} to obtain an image with a clean background, no other objects, and no items inside containers.
All three methods take the regenerated images as input.

\paragraph{Evaluation rules.}
Each trial shows a reference image and three 3D models, whose order is shuffled in every trial.
Participants can rotate a model by dragging, zoom with the scroll wheel, and view its motion from closed to open with a slider or a play button.
Before the study, an instruction page lists the following three questions and their criteria, which are answered in this order in every trial:
\begin{enumerate}
  \item \emph{Motion correctness.} Participants drag the slider to view the opening and closing motion of each model, mark the models whose motion is incorrect, and choose a reason; if all models are correct, they go to the next question. They can go on only after viewing the motion of all three models. The criteria are:
  \begin{itemize}
    \item Joint type: whether the type (rotating, sliding, or detachable) is correct;
    \item Axis: whether the position and direction of the axis are accurate, e.g., a door should rotate about the hinge on its edge;
    \item Moving part: whether the whole part moves, without extra or missing pieces;
    \item Joint count: whether any movable joint is missing or any extra motion appears.
  \end{itemize}
  A model that is not marked as incorrect is counted as moving correctly.
  \item \emph{Fidelity.} Ignoring the joint motion and considering only the structure and texture of the reference image and the 3D models, participants choose the 3D model that is the most similar to the reference image.
  \item \emph{Texture quality.} Regardless of the similarity to the reference image, participants choose the 3D model with the most appealing colors, materials, and patterns.
\end{enumerate}
The fidelity and texture questions provide a ``cannot decide'' option, which is not counted in the vote shares.

\paragraph{Procedure.}
Each participant first completes one practice trial, which is not counted.
Participants are told that the study contains attention checks, but not which trials they are.
In the attention-check trial, one of the three models is deliberately broken by rotating its joint axis by 90$^\circ$, which verifies that participants answer carefully.

\paragraph{Agent-based baselines.}
Both agent-based methods, GPT-6 Astra~\cite{gpt6astra} + Codex~\cite{codex} and GPT-6 Astra + Articraft~\cite{articraft}, use the same prompt, ``Please reconstruct the articulated object in the image. The articulation structure should be correct, realistic, and visually appealing.'', together with the reference image as input.
We do not limit the number of requests, and both methods successfully generate results for all samples.

\paragraph{Statistics.}
Table~\ref{tab:study_ci} summarizes the statistics of the study.
The 95\% confidence intervals are obtained by resampling participants with replacement 10,000 times and recomputing the metrics each time.
All 30 participants mark the broken model in the attention-check trial as incorrect.

\begin{table}[!htbp]
\centering
\footnotesize
\caption{Statistics of the user study (\%). Codex and Articraft denote GPT-6 Astra + Codex and GPT-6 Astra + Articraft. Brackets: 95\% confidence intervals. Agreement: share of participant pairs that choose the same method for the same sample. $p$ vs.\ chance: binomial test of the vote share of Ours against the chance level of 1/3.}
\label{tab:study_ci}
\begin{tabular*}{\linewidth}{@{\extracolsep{\fill}}lccc@{}}
\toprule
 & Motion & Fidelity & Texture \\
\midrule
Codex & 87.7\,{\scriptsize\color{black!70}[85.2, 90.1]} & 4.8\,{\scriptsize\color{black!70}[3.1, 6.9]} & 4.8\,{\scriptsize\color{black!70}[2.9, 7.0]} \\
Articraft & 88.8\,{\scriptsize\color{black!70}[85.6, 91.5]} & 18.3\,{\scriptsize\color{black!70}[13.2, 23.9]} & 15.5\,{\scriptsize\color{black!70}[10.1, 21.5]} \\
\textbf{Ours} & 86.1\,{\scriptsize\color{black!70}[82.3, 89.4]} & \textbf{76.9}\,{\scriptsize\color{black!70}[71.0, 82.4]} & \textbf{79.7}\,{\scriptsize\color{black!70}[72.9, 85.6]} \\
\midrule
Agreement & -- & 67.9 & 72.2 \\
$p$ vs.\ chance & -- & $<10^{-100}$ & $<10^{-100}$ \\
\bottomrule
\end{tabular*}
\end{table}

\subsection{Articulation Learning}
\label{sec:detail_learning}

\paragraph{Data processing.}
All data are processed with the official preprocessing scripts of Particulate~\cite{particulate}.
For PartNet-Mobility~\cite{sapien}, we use the official processed list of Particulate.
For Articraft-10K~\cite{articraft}, we exclude the assets that the scripts do not support, mostly those with more than 16 parts, leaving 9,854 assets.
For \data{}, we remove all objects with free joints and, from the remaining assets, draw 9,854 assets, the same number as for Articraft-10K, with each category receiving a quota proportional to its share.

\paragraph{Training and evaluation.}
All training parameters and evaluation metrics follow the official definitions and code of Particulate.

\paragraph{Per-category results.}
Table~\ref{tab:learning_cat} reports all six metrics on each of the 16 categories of ACD-135 for the three training sets.
Training on \data{} gives the best rest-state gIoU, open-state gIoU, and rest-state mIoU on 14 of the 16 categories each, and the best rest-state PC on 12 categories, showing that its advantage holds across most categories rather than coming from a few of them.

\begin{table*}[t]
\centering
\footnotesize
\setlength{\tabcolsep}{2.5pt}
\caption{Per-category results of Particulate on ACD-135 with different training data. PM: PartNet-Mobility. AC: Articraft-10K. Ours: \data{}. The number of test objects of each category is in parentheses, and leading zeros are omitted. The best result of each category is in bold. The ``All'' row is the same as Table~\ref{tab:preliminary_learning}.}
\label{tab:learning_cat}
\begin{tabular}{@{}l@{\hspace{6pt}}ccc@{\hspace{7pt}}ccc@{\hspace{7pt}}ccc@{\hspace{7pt}}ccc@{\hspace{7pt}}ccc@{\hspace{7pt}}ccc@{}}
\toprule
\multirow{2}{*}{Category} & \multicolumn{3}{c}{Rest gIoU $\uparrow$} & \multicolumn{3}{c}{Rest PC $\downarrow$} & \multicolumn{3}{c}{Rest mIoU $\uparrow$} & \multicolumn{3}{c}{Open gIoU $\uparrow$} & \multicolumn{3}{c}{Open PC $\downarrow$} & \multicolumn{3}{c}{Open OC $\downarrow$} \\
\cmidrule(lr){2-4} \cmidrule(lr){5-7} \cmidrule(lr){8-10} \cmidrule(lr){11-13} \cmidrule(lr){14-16} \cmidrule(lr){17-19}
 & PM & AC & Ours & PM & AC & Ours & PM & AC & Ours & PM & AC & Ours & PM & AC & Ours & PM & AC & Ours \\
\midrule
armoire (19) & .334 & .170 & \textbf{.360} & \textbf{.093} & .103 & .096 & .491 & .375 & \textbf{.519} & .329 & .165 & \textbf{.348} & .221 & .236 & \textbf{.177} & .0128 & .0139 & \textbf{.0069} \\
chest of drawers (16) & .349 & .253 & \textbf{.482} & .066 & .084 & \textbf{.052} & .455 & .442 & \textbf{.575} & .322 & .234 & \textbf{.466} & \textbf{.078} & .087 & .083 & .0053 & \textbf{.0020} & .0024 \\
desk (12) & .341 & .116 & \textbf{.353} & \textbf{.073} & .099 & .105 & .453 & .287 & \textbf{.554} & .323 & .110 & \textbf{.337} & \textbf{.091} & .116 & .137 & .0029 & .0026 & \textbf{.0012} \\
nightstand (11) & .362 & .271 & \textbf{.756} & .117 & .151 & \textbf{.016} & .506 & .458 & \textbf{.772} & .346 & .269 & \textbf{.730} & .161 & .188 & \textbf{.017} & .0114 & .0116 & \textbf{.0013} \\
bookcase (10) & .167 & .065 & \textbf{.263} & \textbf{.079} & .128 & .107 & .342 & .288 & \textbf{.425} & .158 & .062 & \textbf{.252} & \textbf{.116} & .162 & .134 & .0032 & .0029 & \textbf{.0017} \\
oven (9) & .030 & -.092 & \textbf{.266} & .232 & .245 & \textbf{.130} & .347 & .301 & \textbf{.512} & .027 & -.097 & \textbf{.254} & .287 & .328 & \textbf{.263} & \textbf{.0057} & .0066 & .0091 \\
sink cabinet (9) & .266 & .049 & \textbf{.391} & .093 & .130 & \textbf{.084} & .411 & .295 & \textbf{.552} & .244 & .042 & \textbf{.366} & .170 & .210 & \textbf{.150} & .0077 & .0073 & \textbf{.0061} \\
washer (8) & .158 & .193 & \textbf{.808} & .224 & .205 & \textbf{.000} & .458 & .477 & \textbf{.808} & .151 & .191 & \textbf{.805} & .339 & .319 & \textbf{.181} & \textbf{.0046} & .0088 & .0142 \\
hanging cabinet (7) & .569 & .405 & \textbf{.598} & .059 & .081 & \textbf{.033} & \textbf{.667} & .559 & .652 & .552 & .395 & \textbf{.594} & .210 & .244 & \textbf{.107} & .0120 & .0128 & \textbf{.0038} \\
TV stand (6) & .224 & .181 & \textbf{.515} & .114 & .133 & \textbf{.046} & .392 & .399 & \textbf{.597} & .210 & .176 & \textbf{.487} & .141 & .178 & \textbf{.051} & .0015 & .0030 & \textbf{.0005} \\
kitchen cabinet (6) & \textbf{.261} & .171 & .223 & \textbf{.087} & .106 & .108 & .421 & .380 & \textbf{.456} & \textbf{.254} & .166 & .213 & \textbf{.120} & .169 & .204 & .0084 & \textbf{.0061} & .0070 \\
microwave (6) & .435 & .242 & \textbf{.668} & .116 & .154 & \textbf{.025} & .609 & .464 & \textbf{.707} & .431 & .239 & \textbf{.661} & \textbf{.143} & .338 & .196 & \textbf{.0098} & .0159 & .0142 \\
table (6) & \textbf{.238} & .147 & .204 & .133 & .124 & \textbf{.111} & \textbf{.449} & .332 & .407 & \textbf{.219} & .141 & .196 & .137 & .141 & \textbf{.128} & \textbf{.0017} & .0019 & .0021 \\
refrigerator (5) & .275 & .149 & \textbf{.409} & .113 & .138 & \textbf{.082} & .501 & .377 & \textbf{.542} & .267 & .146 & \textbf{.405} & .178 & .179 & \textbf{.140} & .0078 & \textbf{.0054} & .0099 \\
cabinet (3) & .177 & .258 & \textbf{.416} & .127 & .127 & \textbf{.057} & .425 & .445 & \textbf{.509} & .169 & .240 & \textbf{.402} & \textbf{.129} & .144 & .400 & \textbf{.0055} & .0061 & .0293 \\
dishwasher (2) & .665 & .385 & \textbf{.900} & .062 & .193 & \textbf{.004} & .749 & .656 & \textbf{.900} & .664 & .385 & \textbf{.900} & \textbf{.072} & .299 & \textbf{.072} & \textbf{.0081} & .0280 & .0087 \\
\midrule
All (135) & .295 & .171 & \textbf{.452} & .109 & .130 & \textbf{.071} & .464 & .390 & \textbf{.578} & .282 & .164 & \textbf{.439} & .166 & .199 & \textbf{.143} & .0071 & .0076 & \textbf{.0059} \\
\bottomrule
\end{tabular}
\end{table*}

\section{Additional Experiments}
\label{sec:additional}

This section reports two additional experiments on ACD-135~\cite{s2o} under the same protocol as Sec.~\ref{sec:compare_single_image}:
a comparison with FreeArt3D~\cite{freeart3d} (Sec.~\ref{sec:cmp_prior})
and an ablation study of \ours{} (Sec.~\ref{sec:ablation}).
Both use \ours{} with the same settings as in the main text.

\subsection{Comparison with FreeArt3D}
\label{sec:cmp_prior}

Besides FreeArt3D~\cite{freeart3d}, DreamArt~\cite{dreamart} and Rest2Art~\cite{rest2art}, discussed in Sec.~\ref{sec:related}, also use pretrained generative models to recover articulation.
However, neither of them has released its code at the time of writing, so we cannot run them under the same setting.
We therefore compare \ours{} with FreeArt3D.

\paragraph{Setup.}
FreeArt3D takes as input a set of images of the same object in several motion states.
Its official evaluation code renders these images directly from the ground-truth model:
six states evenly spaced from closed to fully open within the joint range, with a disk at the same position under the object in every state.
The disk is needed because FreeArt3D reconstructs each state separately and does not estimate the relative poses between the states.
Since the disk appears at the same position in every image, it serves as a shared static reference for aligning all states to one coordinate frame.

We prepare the inputs of FreeArt3D following its official code.
We render six motion states from the ground-truth models of ACD-135 and add the disk with the official code.
For objects with multiple joints, we follow the official multi-joint code: we render one set of states for each joint with the other joints closed, plus one image with all joints open.
Also following the official code, the number and types of joints are given by the ground truth.
In contrast, \ours{} takes only a single open-state image as input, and the number and types of joints are recognized by the VLM.

FreeArt3D outputs one mesh for each state.
We use the mesh of the fully open state and the predicted joints as its result.
As with \ours{}, we compute its metrics in the fully open state under the SINGAPO protocol.

\paragraph{Results.}
FreeArt3D uses extra information that is not available in practice:
ground-truth images of multiple states, the ground-truth number and types of joints, and the disk for aligning the states.
\ours{} uses only a single image.
Even so, \ours{} clearly outperforms FreeArt3D on both all objects and the single-joint subset (Table~\ref{tab:cmp_prior}).
On all 135 objects, \ours{} is better on all five metrics.
Its RS-$d_\text{gIoU}$ is 0.517 vs.\ 0.802, about 36\% lower;
its RS-$d_\text{cDist}$ is 0.060 vs.\ 0.172, about 65\% lower;
and its AS-$d_\text{cDist}$ and AOR are about 59\% and 78\% lower.
Per object, \ours{} has a lower RS-$d_\text{gIoU}$ on 116 of the 135 objects.

\begin{table}[!htbp]
\centering
\small
\caption{Comparison with FreeArt3D. Lower is better for all metrics.
The numbers of \ours{} on all 135 objects are the same as in Table~\ref{tab:method_acd}.
AOR is not reported for single-joint objects, since it needs at least two movable parts.}
\label{tab:cmp_prior}
\begin{tabular*}{\columnwidth}{@{\extracolsep{\fill}}lccccc@{}}
\toprule
\multirow{2}{*}{Method} & \multicolumn{2}{c}{$d_\text{gIoU}$} & \multicolumn{2}{c}{$d_\text{cDist}$} & \multirow{2}{*}{AOR} \\
\cmidrule(lr){2-3}\cmidrule(lr){4-5}
 & RS & AS & RS & AS & \\
\midrule
\multicolumn{6}{@{}l}{\emph{All objects (135)}} \\
FreeArt3D~\cite{freeart3d} & 0.802 & 0.803 & 0.172 & 0.174 & 0.0185 \\
\ours{} (ours) & \textbf{0.517} & \textbf{0.525} & \textbf{0.060} & \textbf{0.072} & \textbf{0.0041} \\
\midrule
\multicolumn{6}{@{}l}{\emph{Single-joint objects (25)}} \\
FreeArt3D~\cite{freeart3d} & 0.527 & 0.526 & 0.134 & 0.129 & -- \\
\ours{} (ours) & \textbf{0.350} & \textbf{0.354} & \textbf{0.044} & \textbf{0.050} & -- \\
\bottomrule
\end{tabular*}
\end{table}

We also report the 25 objects in ACD-135 with a single movable joint, which is the setting targeted by the test set of FreeArt3D.
On this subset, the RS-$d_\text{gIoU}$ of \ours{} is 0.350 vs.\ 0.527, about 34\% lower,
and its RS-$d_\text{cDist}$ is 0.044 vs.\ 0.134, about 67\% lower.

\begin{figure}[!htbp]
\centering
\includegraphics[width=\columnwidth]{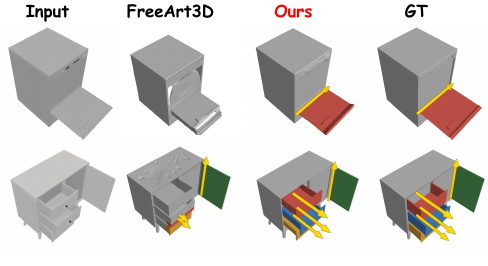}
\caption{Qualitative comparison with FreeArt3D~\cite{freeart3d} on a single-joint object (top) and a multi-joint object (bottom) of ACD-135, in the open state.
Each movable part takes the color of the ground-truth part it is matched to, and the base is shown in gray.}
\label{fig:cmp_prior}
\end{figure}

The more joints an object has, the larger the gap.
When we group the objects by the number of ground-truth joints, the RS-$d_\text{gIoU}$ gap between the two methods grows from 0.177 for one joint (0.350 vs.\ 0.527)
to 0.225 for two or three joints (0.503 vs.\ 0.728), and to 0.411 for four or more joints (0.618 vs.\ 1.029).
This shows that the advantage of \ours{} is larger on objects with more complex structures.
Qualitative results are shown in Fig.~\ref{fig:cmp_prior}.

\subsection{Ablation Study}
\label{sec:ablation}

As referenced in Sec.~\ref{sec:compare_single_image}, we ablate the key components of \ours{} on ACD-135, under the same evaluation protocol as the main text.
Each variant replaces only one component and keeps all other settings unchanged.
All variants reuse the motion videos generated for the main-paper experiments, so the differences come only from the ablated component.
Table~\ref{tab:ablation} summarizes the results.

\begin{table*}[t]
\centering
\small
\caption{Ablation study of \ours{} on ACD-135. Lower is better for all metrics.}
\label{tab:ablation}
\begin{tabular*}{\textwidth}{@{\extracolsep{\fill}}llccccc@{}}
\toprule
\multirow{2}{*}{Component} & \multirow{2}{*}{Variant} & \multicolumn{2}{c}{$d_\text{gIoU}$} & \multicolumn{2}{c}{$d_\text{cDist}$} & \multirow{2}{*}{AOR} \\
\cmidrule(lr){3-4}\cmidrule(lr){5-6}
 & & RS & AS & RS & AS & \\
\midrule
-- & Full method & 0.517 & 0.525 & 0.060 & 0.072 & 0.0041 \\
\midrule
Keyframe selection & Uniform in time & 0.527 & 0.534 & 0.060 & 0.074 & 0.0051 \\
\midrule
\multirow{2}{*}{State alignment} & No alignment (CUPID poses) & 0.894 & 0.901 & 0.235 & 0.252 & 0.0090 \\
 & Least squares, w/o RANSAC & 0.716 & 0.720 & 0.140 & 0.153 & 0.0074 \\
\midrule
\multirow{2}{*}{Axis fitting} & Tracking only & 0.515 & 0.522 & 0.059 & 0.072 & 0.0044 \\
 & Plane intersection only & 0.634 & 0.641 & 0.128 & 0.137 & 0.0059 \\
\midrule
Interior completion & w/o interior completion & 0.605 & 0.610 & 0.081 & 0.090 & 0.0057 \\
\bottomrule
\end{tabular*}
\end{table*}

\begin{figure*}[tp]
\centering
\includegraphics[width=\textwidth]{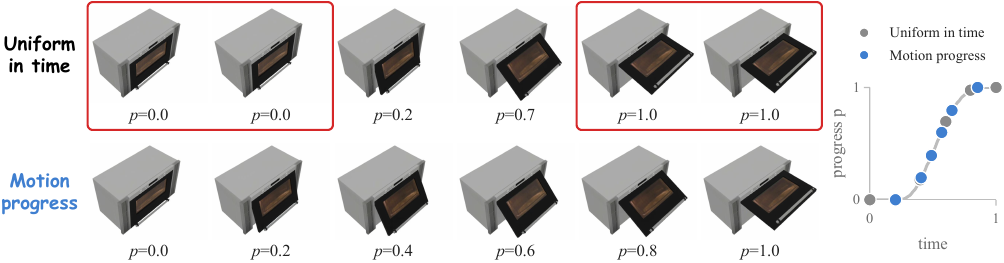}
\caption{Comparison of sampling keyframes uniformly in time and by motion progress.
Red boxes mark the repeated closed and fully open frames of uniform sampling.}
\label{fig:abl_keyframes}
\end{figure*}

\paragraph{Keyframe selection.}
We replace keyframe selection by motion progress (Sec.~\ref{sec:keyframe}) with frames sampled uniformly in time over the whole video.
RS-$d_\text{gIoU}$ increases from 0.517 to 0.527, and AS-$d_\text{gIoU}$ and AS-$d_\text{cDist}$ also become worse.
In some generated videos, the motion speed is not uniform: the part barely moves at first and then opens quickly.
With uniform sampling in time, most keyframes fall in the slow stages, where adjacent states are almost the same, while the fast stage is skipped.
As a result, cross-state voting and axis fitting get less useful motion information.
Sampling by motion progress instead makes every two adjacent states differ by a clear motion.
Fig.~\ref{fig:abl_keyframes} compares the two sampling methods.

\begin{figure}[!htbp]
\centering
\includegraphics[width=\columnwidth]{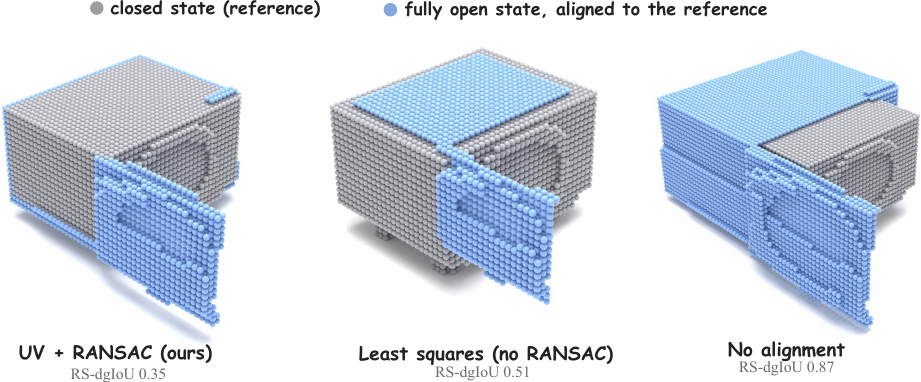}
\caption{Comparison of the three state alignment methods.
Gray: the closed state (reference). Blue: the voxels of the fully open state after alignment.}
\label{fig:abl_alignment}
\end{figure}

\paragraph{State alignment.}
We test two variants.
The first does not align the states and directly uses the poses predicted by CUPID.
The second still aligns the states, but estimates the similarity transform from all UV correspondences by least squares, without RANSAC.
Alignment is the component with the largest effect.
Without alignment, RS-$d_\text{gIoU}$ increases from 0.517 to 0.894, and RS-$d_\text{cDist}$ increases from 0.060 to 0.235, about four times larger.
CUPID reconstructs each state independently, so the scale, position, and even the $90^\circ$ orientation differ between states.
Without alignment, the static base does not overlap across states.
Cross-state voting then mistakes it for a moving part, and the following part segmentation and axis fitting both fail.
With least squares, RS-$d_\text{gIoU}$ is 0.716, still clearly worse than the full method.
The reason is that some correspondences lie on the moving parts and pull the least-squares fit away.
RANSAC keeps only the consistent correspondences on the base, so the moving parts do not affect the alignment.
Fig.~\ref{fig:abl_alignment} compares the three alignment methods.

\paragraph{Axis fitting.}
We replace the automatic choice (Sec.~\ref{sec:joint_fitting}) by using only tracking, or only plane intersection, for all doors.
With plane intersection only, RS-$d_\text{gIoU}$ increases from 0.517 to 0.634.
Plane intersection requires the part to be a flat panel whose normal changes clearly during the rotation.
For doors that do not meet this condition, plane intersection cannot give an axis.
Examples are curved doors, and doors whose planes are almost parallel across states, so that no valid intersection can be found.
In this setting, no axis can be fitted for any door in 9 samples, so no articulated object is output for them, and they are counted as failures with the worst values (Sec.~\ref{sec:detail_motion}).
This causes the increase in the metrics.
With tracking only, all metrics are close to those of the full method, because tracking works on all the samples that plane intersection can handle.

\paragraph{Interior completion.}
We remove interior completion (Sec.~\ref{sec:completion_generation}):
the drawer boxes are not completed, only the segmented drawer fronts are kept, and the sliding range stays the same.
RS-$d_\text{gIoU}$ increases from 0.517 to 0.605, and RS-$d_\text{cDist}$ from 0.060 to 0.081.
In the generated open states, only the drawer fronts are visible, and the inside of the base is often reconstructed as solid.
Without interior completion, a predicted drawer is only a part of its front, which is very different from the real drawer box.
With completion, each drawer has a complete box and the base has the matching cavity, so the part geometry is consistent with the real structure.

\section{More Details on Dataset Construction}
\label{sec:construction_details}

This section provides more details on the human inspection (Sec.~\ref{sec:inspection}) and the physical property estimation (Sec.~\ref{sec:phys_annot}) of \data{} (Sec.~\ref{sec:dataset}).

\subsection{Human Inspection}
\label{sec:inspection}

Human inspection is carried out by 10 teams of five people, each consisting of one inspector and four annotators.
It has three stages:
\begin{itemize}
  \item \emph{Stage 1} works on the generated sparse voxels: annotators first filter out samples with poor generation quality and then correct the structural errors of revolute joints.
  \item \emph{Stage 2}: after the samples kept in Stage 1 are decoded into meshes, inspectors first review the results of the annotators; the approved samples then have their joint motion positions refined and their texture quality further screened.
  \item \emph{Stage 3}: the data completed in Stage 2 are accepted through batch-wise spot checks, and only accepted data enter the final dataset.
\end{itemize}
All inspection steps are performed on an in-house data inspection platform with a graphical user interface.
As stated in Sec.~\ref{sec:dataset}, samples with defects in the generated geometry, interior structure, or textures are discarded directly by the inspectors and are never repaired.
Manual edits are limited to refining joint parameters: the axes and part segmentation of revolute joints (Stage 1) and the closed positions of movable parts (Stage 2).

\paragraph{Stage 1: voxel stage (annotators).}
\emph{Filtering.}
This step removes unusable samples as early as possible, so that no correction effort is wasted on them.
Annotators check the overall generation quality of each sample.
The main reasons for rejection include:
\begin{itemize}
  \item obvious errors in the axes or part segmentation;
  \item implausible interior structures;
  \item incorrect numbers or types of joints;
  \item floating debris, or debris on invisible regions such as the back of the object;
  \item incorrect overall geometry;
  \item objects other than the target articulated object in the generated voxels.
\end{itemize}
Samples that fail are filtered out in this step.

\emph{Refining joint axes and part segmentation.}
This step refines the joint parameters of the samples that pass the filtering.
In this step, annotators only correct revolute joints and leave prismatic and free joints unchanged: they check the position and orientation of each axis and the accuracy of the part segmentation, and manually adjust inaccurate axes and segmentations.

\paragraph{Stage 2: mesh stage (inspectors).}
\emph{Reviewing the voxel-stage results.}
This step checks the work of the annotators and sends problems back to their source.
Inspectors first screen the overall quality of the decoded meshes.
For a sample that fails, the inspector reviews the voxel-stage result of the annotator, identifies the problem, and writes down the reason.
If the problem lies in the joint axes or the part segmentation, the sample is returned to the annotator for correction; otherwise, it is discarded.
Only approved samples proceed to the next two steps.

\emph{Refining motion positions.}
Because the resolution of sparse voxels is low, the closed positions of movable parts obtained in Stage 1 often have small errors, which appear on the mesh as parts that are not fully closed or that penetrate the base.
Inspectors therefore check the full motion of every joint and correct the closed positions of revolute and prismatic joints.

\emph{Texture selection and screening.}
To obtain cleaner and more appealing textures, we generate four textures for each object with four random seeds, and the inspector selects the cleanest and most appealing one.
If all textures or the overall appearance are poor, the sample is discarded.
The data completed by the inspectors then enter the acceptance stage.

\paragraph{Stage 3: acceptance.}
The data approved by each inspector are grouped into batches of 400 and spot-checked batch by batch: 10\% of each batch is sampled, and batches with a pass rate of at least 95\% enter the accepted pool.
A batch that fails is returned to the inspector as a whole for self-checking: problematic samples whose joint parameters are wrong are corrected as in Stages 1 and 2, and the other problematic samples are discarded.
The batch is then spot-checked again, and this process repeats until the batch passes.

Table~\ref{tab:qc_groups} reports the acceptance results of each team.
The average pass rate of the first spot checks is 82.1\%, and after the corrections, the average pass rate at the end of acceptance is 96.4\%.

\begin{table}[!htbp]
\centering
\small
\caption{Acceptance results of each team. First: average pass rate of the first spot check of each batch. Final: average pass rate when all acceptance is finished.}
\label{tab:qc_groups}
\begin{tabular*}{\linewidth}{@{\extracolsep{\fill}}lccc@{}}
\toprule
Team & Batches & First (\%) & Final (\%) \\
\midrule
Team 1 & 28 & 80.2 & 97.7 \\
Team 2 & 30 & 79.8 & 95.4 \\
Team 3 & 29 & 81.3 & 97.3 \\
Team 4 & 29 & 81.0 & 96.6 \\
Team 5 & 29 & 77.9 & 97.2 \\
Team 6 & 31 & 84.6 & 95.8 \\
Team 7 & 30 & 78.7 & 96.3 \\
Team 8 & 33 & 86.2 & 97.1 \\
Team 9 & 30 & 80.2 & 95.6 \\
Team 10 & 29 & 91.0 & 95.4 \\
\midrule
Total & 298 & 82.1 & 96.4 \\
\bottomrule
\end{tabular*}
\end{table}

\subsection{Physical Property Estimation}
\label{sec:phys_annot}

Generated assets only have geometry and textures; to be used in simulation, they also need physical properties such as scale, mass, and friction.
We estimate these properties automatically for every asset in four steps:
we first render whole-object views and part-highlight views;
GPT-6 Astra~\cite{gpt6astra} then estimates the object scale and the physical properties of each part and joint from the renderings;
the estimates then pass two rounds of checks, per asset and per category, and failed ones are re-estimated with their specific problems;
finally, we build collision geometry for each part, compute the centers of mass and inertia, and export the assets to formats that simulators can load directly.
We describe each step below.

\paragraph{Rendering.}
Each object is rendered with all parts closed by Blender Cycles with 32 samples per pixel, denoising, and a resolution of $600 \times 600$.
There are two kinds of renderings:
\begin{itemize}
  \item \emph{Whole-object views}: four views, one from the top and three from the sides at an elevation of $30^\circ$. Each view overlays the axis-aligned bounding box of the object and the RGB coordinate axes, with a caption that gives the aspect ratio of the bounding box and the axis of its longest edge.
  \item \emph{Part-highlight views}: one view per part, from the camera that sees the part best: a drawer is viewed from the side it slides out to, and other parts from the front. The target part keeps its original texture, and the other parts are rendered in gray.
\end{itemize}

\paragraph{Estimated properties.}
The whole-object views, the part-highlight views, the category name, the part list, and the annotated joints are given to GPT-6 Astra, which estimates the properties in Table~\ref{tab:phys_props}.
Centers of mass and inertia are not estimated by the VLM but computed from the mesh and the part masses.
Materials are chosen from 12 categories (Table~\ref{tab:material}).

\begin{table}[!htbp]
\centering
\small
\caption{Physical properties estimated by the VLM. The real-world scale is given by the real length of the longest edge of the bounding box.}
\label{tab:phys_props}
\begin{tabular*}{\linewidth}{@{\extracolsep{\fill}}lll@{}}
\toprule
Level & Property & Type / unit \\
\midrule
Object & Real-world scale & cm \\
\cmidrule{1-3}
\multirow{6}{*}{Part} & Name & text \\
 & Function description & text \\
 & Material & category (Table~\ref{tab:material}) \\
 & Mass & kg \\
 & Static / dynamic friction & coefficient \\
 & Restitution & coefficient \\
\cmidrule{1-3}
\multirow{4}{*}{Joint} & Damping (revolute) & N\,m\,s/rad \\
 & Damping (prismatic) & N\,s/m \\
 & Friction (revolute) & N\,m \\
 & Friction (prismatic) & N \\
\bottomrule
\end{tabular*}
\end{table}

\begin{table}[!htbp]
\centering
\small
\caption{Material categories and their prior ranges of density and static friction, used in the per-asset checks.}
\label{tab:material}
\begin{tabular*}{\linewidth}{@{\extracolsep{\fill}}lcc@{}}
\toprule
Material & Density (kg/m$^3$) & Friction \\
\midrule
Metal & 2600--8960 & 0.15--0.8 \\
Plastic & 850--1500 & 0.15--0.7 \\
Fabric & 100--1500 & 0.3--1.2 \\
Leather & 600--1100 & 0.4--1.0 \\
Foam & 15--300 & 0.3--1.0 \\
Rubber & 900--1500 & 0.5--1.6 \\
Glass & 2200--2800 & 0.2--0.9 \\
Stone & 2200--3100 & 0.4--1.2 \\
Ceramics & 1800--6000 & 0.3--1.0 \\
Wood & 300--1100 & 0.3--0.9 \\
Concrete & 1800--2600 & 0.5--1.1 \\
Other & 50--3000 & 0.1--1.2 \\
\bottomrule
\end{tabular*}
\end{table}

\paragraph{First round: per-asset checks.}
Each estimate is checked as soon as it is returned, and clearly impossible or invalid values are marked as errors that require re-estimation:
\begin{itemize}
  \item \emph{Value ranges}: the longest edge is outside 1--600\,cm; the density is outside 0.01--22\,g/cm$^3$; a friction coefficient is outside 0.02--2, or the dynamic friction exceeds the static friction; the restitution coefficient is outside 0--1; or the material category is unknown.
  \item \emph{Convex-hull-based mass bounds}: let $V$ and $A$ be the convex-hull volume and the surface area of a part, and $[\rho_{\min}, \rho_{\max}]$ the density range of its material (Table~\ref{tab:material}). The mass $m$ of the part must satisfy
  \begin{equation}
    0.8\, \rho_{\min}\, \frac{A}{2}\, t \;\le\; m \;\le\; 1.2\, \rho_{\max}\, V, \qquad t = 0.3\,\text{mm},
    \label{eq:mass_bounds}
  \end{equation}
  that is, a part can be neither heavier than its convex hull filled with the densest version of its material, nor lighter than a 0.3\,mm shell over half of its surface area.
\end{itemize}

\paragraph{Second round: per-category checks.}
After all objects are estimated, we compute, within each category, the robust z-scores of the logarithms of size, mass, and density (mass divided by convex-hull volume):
\begin{equation}
  z = \frac{\log x - \operatorname{median}}{\max\left(1.4826 \cdot \operatorname{MAD},\ \text{floor}\right)},
  \label{eq:robust_z}
\end{equation}
where the median and the median absolute deviation (MAD) are computed within the category, and the floor prevents a tiny denominator when the values in a category are concentrated.
Objects in the same category can differ greatly; for example, both a small desktop shelf and a two-meter bookcase are bookcases.
We therefore re-estimate only clear outliers, whose $|z| > 3$ and whose values also deviate from the category median by a large factor (4$\times$ for size and 8$\times$ for mass and density).

\paragraph{Re-estimation.}
A failed estimate is re-queried with the same images and prompt, together with the previous answer and a textual description of its specific problems.
The description contains no bounds, category medians, or values of other samples; it only asks the model to re-check and to keep a value if the images clearly support it.
The new answer is adopted only if it has fewer problems.
Objects that deviate only in the per-category statistics and keep their values after re-checking retain their original estimates.

\paragraph{Collision geometry, inertia, and export.}
\begin{itemize}
  \item \emph{Collision geometry}: each part is decomposed into convex pieces by CoACD~\cite{coacd}. The input is first simplified to at most 20,000 faces, the threshold is 0.05, and the base and each other part have at most 48 and 16 pieces, respectively. The collision geometry is stored as the second scene of the released \texttt{object.glb}.
  \item \emph{Centers of mass and inertia}: recomputed by distributing the mass of each part uniformly over its convex pieces, and written to \texttt{object.json}.
  \item \emph{Release format}: all assets use meters, $+Z$ as the up axis, the bottom at $z = 0$, and the closed state; the visual meshes are simplified to about 120k faces with re-baked textures.
  \item \emph{Export}: our conversion tool exports the released files to URDF or MJCF in one step, which can be loaded directly into simulators such as MuJoCo~\cite{mujoco}, SAPIEN~\cite{sapien}, Genesis~\cite{genesis}, and Isaac Sim~\cite{isaacsim}.
\end{itemize}

\section{More Dataset Examples}
\label{sec:examples}

\begin{figure*}[p]
\centering
\includegraphics[width=\textwidth]{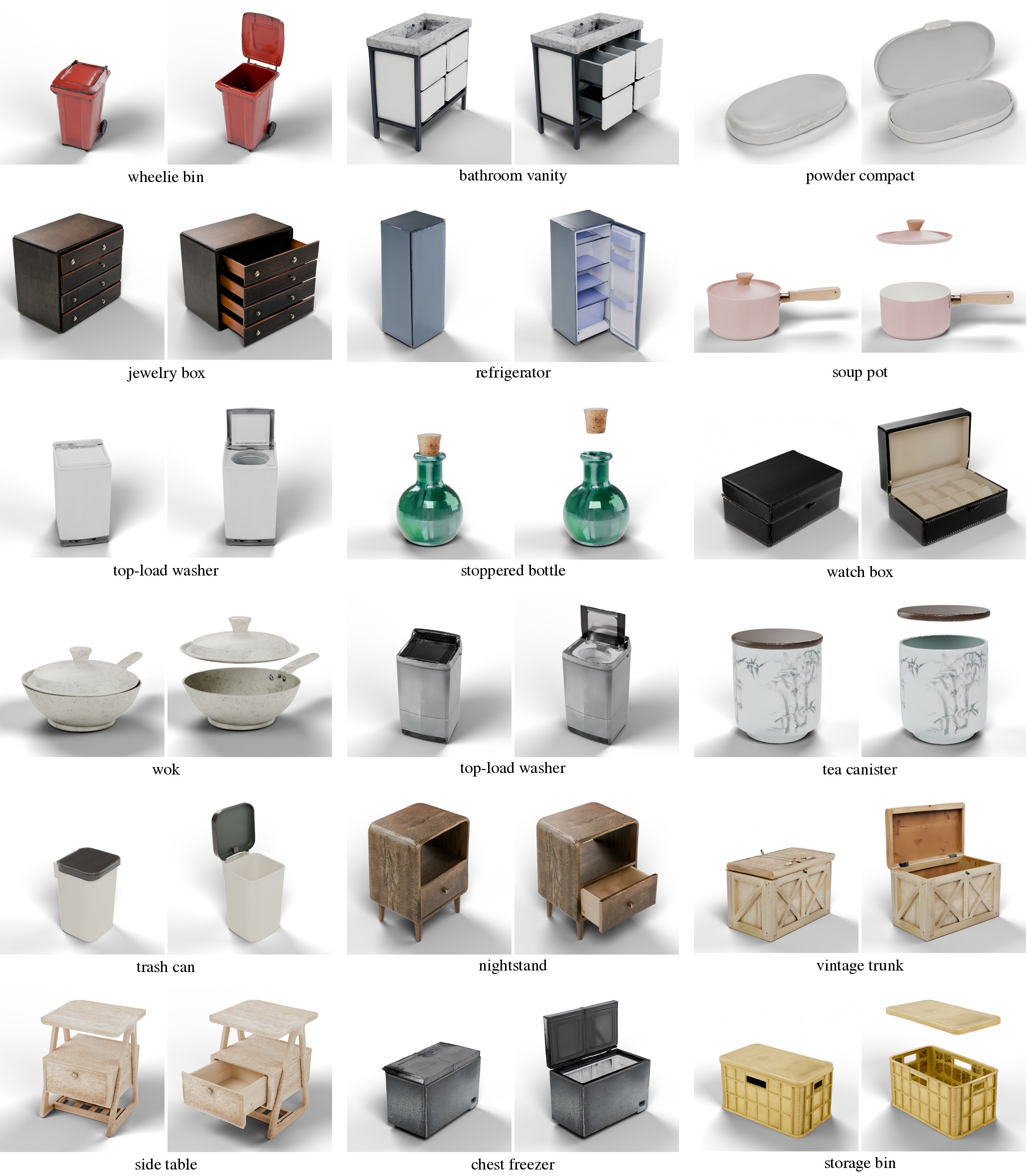}
\caption{Examples of \data{} (1/5). For each asset, the left image shows the closed state and the right image shows the open state.}
\label{fig:examples_1}
\end{figure*}

\begin{figure*}[p]
\centering
\includegraphics[width=\textwidth]{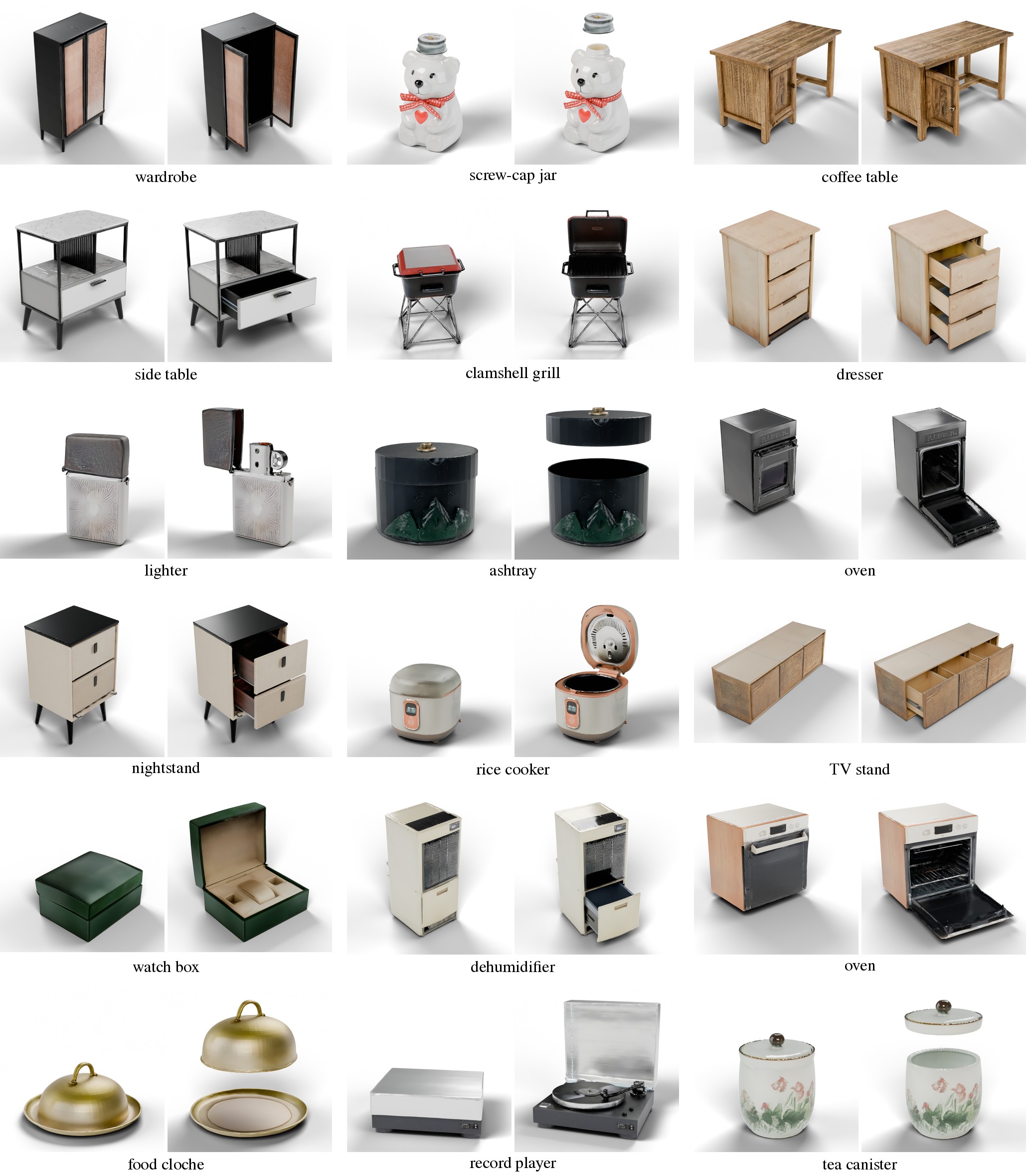}
\caption{Examples of \data{} (2/5). For each asset, the left image shows the closed state and the right image shows the open state.}
\label{fig:examples_2}
\end{figure*}

\begin{figure*}[p]
\centering
\includegraphics[width=\textwidth]{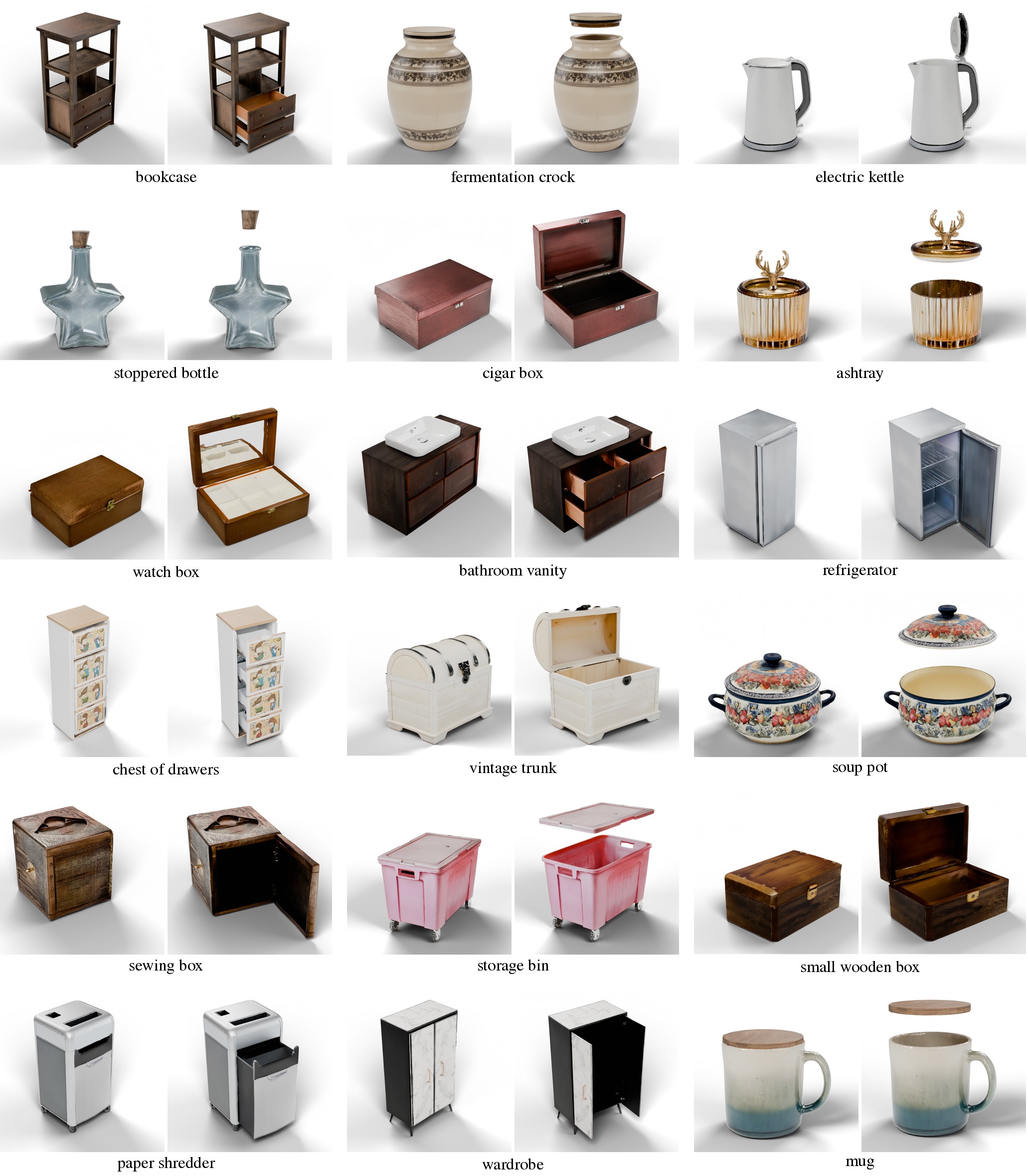}
\caption{Examples of \data{} (3/5). For each asset, the left image shows the closed state and the right image shows the open state.}
\label{fig:examples_3}
\end{figure*}

\begin{figure*}[p]
\centering
\includegraphics[width=\textwidth]{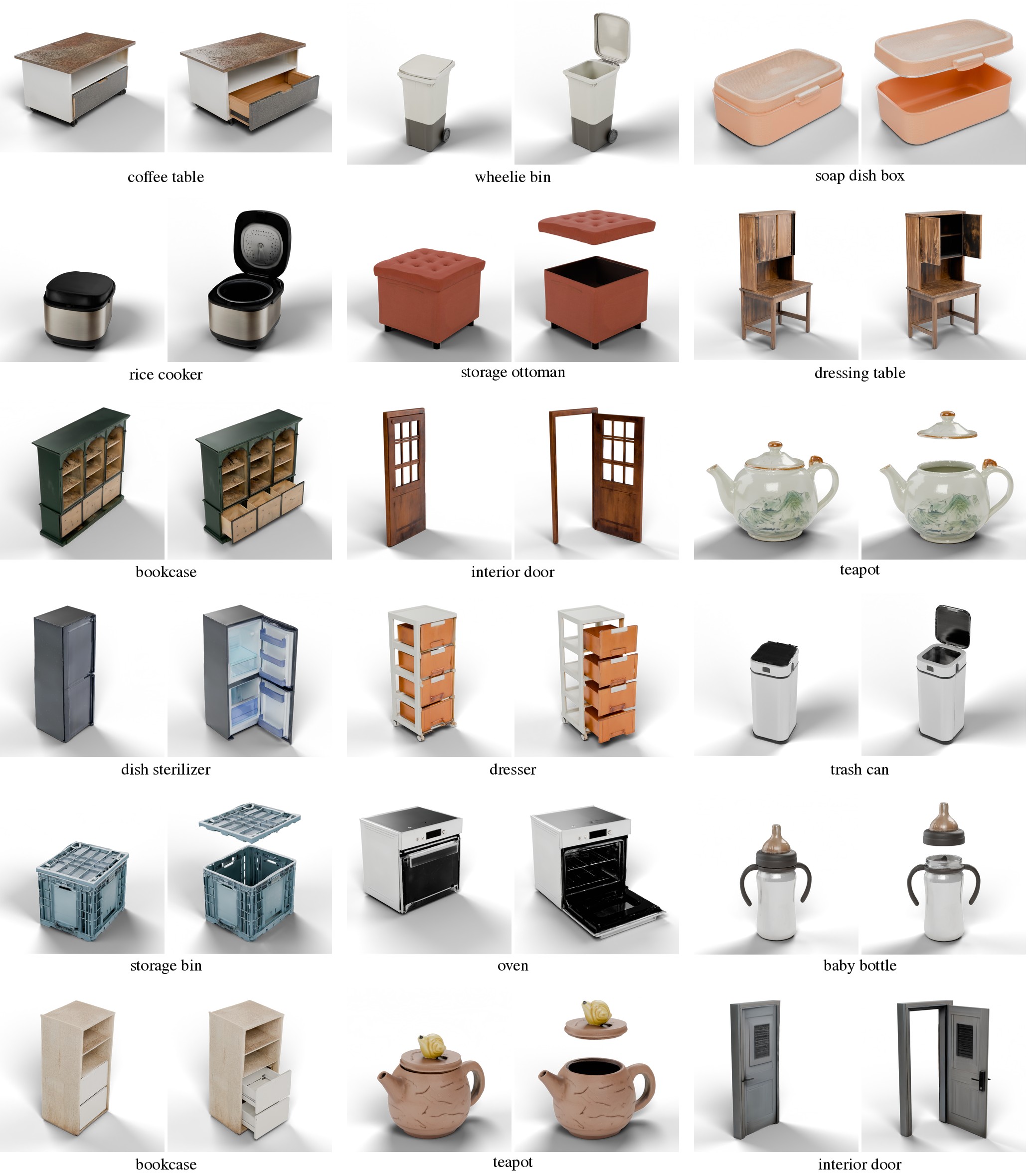}
\caption{Examples of \data{} (4/5). For each asset, the left image shows the closed state and the right image shows the open state.}
\label{fig:examples_4}
\end{figure*}

\begin{figure*}[p]
\centering
\includegraphics[width=\textwidth]{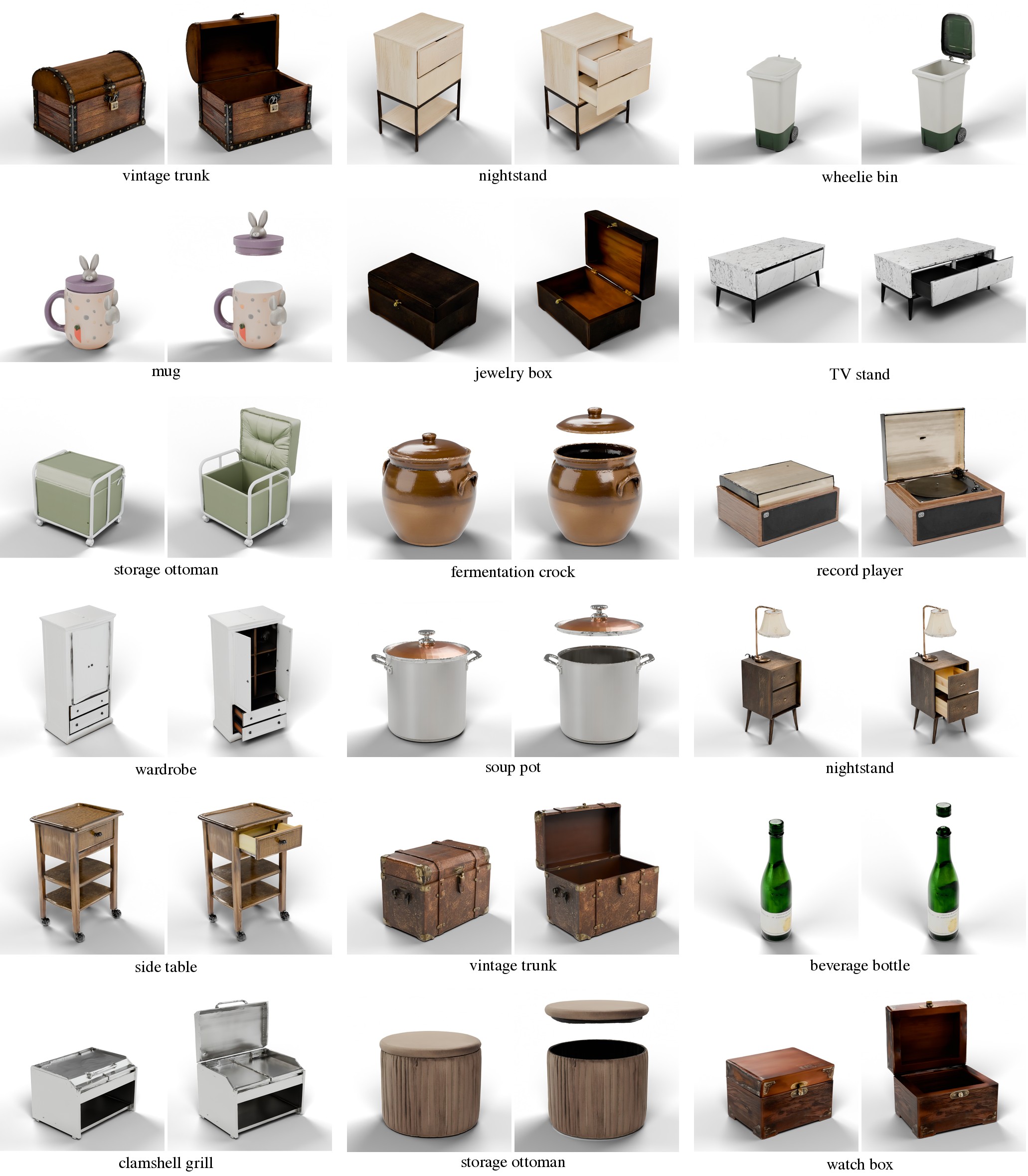}
\caption{Examples of \data{} (5/5). For each asset, the left image shows the closed state and the right image shows the open state.}
\label{fig:examples_5}
\end{figure*}

Figs.~\ref{fig:examples_1}--\ref{fig:examples_5} show 90 assets of \data{}.
For each asset, the left image shows the closed state and the right image shows the open state.

\end{document}